\pdfoutput=1

\documentclass[11pt]{article}
\usepackage[final]{acl}
\usepackage{booktabs}
 \usepackage{amsmath} 
 \usepackage{amssymb}
\usepackage{times}
\usepackage{latexsym}
\usepackage{tabularx}

\usepackage[T1]{fontenc}
\usepackage{symbols}
\usepackage[utf8]{inputenc}

\usepackage{microtype}

\usepackage{inconsolata}

\usepackage{graphicx}
\usepackage{booktabs}
\usepackage{multirow}
\usepackage{array}
\usepackage{caption}
\usepackage{subcaption}

\usepackage{xcolor}
\usepackage[most]{tcolorbox} %

\definecolor{taskbg}{RGB}{248,248,255}
\definecolor{taskframe}{RGB}{80,80,140}

\newtcolorbox{taskbox}{
  enhanced,
  colback=taskbg,
  colframe=taskframe,
  boxrule=0.6pt,
  arc=2mm,
  left=6pt,right=6pt,top=6pt,bottom=6pt,
  before skip=0.6\baselineskip,
  after skip=0.6\baselineskip,
  breakable,
  before upper={\setlength{\parindent}{0pt}},
}

\usepackage{multirow} 
\usepackage{pifont}    %
\usepackage{xcolor}    %
\usepackage[toc,page]{appendix}
\newcommand{\cmark}{\textcolor{green!60!black}{\ding{51}}} %

\usepackage[most]{tcolorbox}

\usepackage[most]{tcolorbox}
\tcbuselibrary{breakable,skins}
\usepackage{fvextra} %

\newenvironment{promptbox}{%
  \VerbatimEnvironment
  \begin{tcolorbox}[
    enhanced, breakable,
    colback=black!2, colframe=black!35,
    boxrule=0.5pt, arc=2pt,
    left=6pt,right=6pt,top=6pt,bottom=6pt,
  ]%
  \begin{Verbatim}[
    fontsize=\footnotesize,
    breaklines=true,
    breakanywhere=true,
    breaksymbolleft={},
    breaksymbolright={}
  ]%
}{%
  \end{Verbatim}%
  \end{tcolorbox}%
}

\title{Can MLLMs Find Their Way in a City? Exploring Emergent Navigation from Web-Scale Knowledge}

\author{Dwip Dalal\textsuperscript{1}\thanks{Equal contribution.} \thanks{Work performed during a research internship at Microsoft Research, Redmond.} \quad
  Utkarsh Mishra\textsuperscript{2}\footnotemark[1] \quad 
  Narendra Ahuja\textsuperscript{1} \quad
  Nebojsa Jojic\textsuperscript{3} \\
  \textsuperscript{1}University of Illinois Urbana-Champaign, 
  \textsuperscript{2}Texas A\&M University \\  
  \textsuperscript{3}Microsoft Research, Redmond \\ 
  \texttt{dwip2@illinois.edu}
}

\begin{document}
\maketitle
\begin{abstract}
\vspace*{-12pt}
Leveraging multimodal large language models (MLLMs) to develop embodied agents offers significant promise for addressing complex real-world tasks. However, current evaluation benchmarks remain predominantly language-centric or heavily reliant on simulated environments, rarely probing the nuanced, knowledge-intensive reasoning essential for practical, real-world scenarios. To bridge this critical gap, we introduce the task of \emph{Sparsely Grounded Visual Navigation}, explicitly designed to evaluate the sequential decision-making abilities of MLLMs in challenging, knowledge-intensive real-world environment. We operationalize this task with \ourmethodshort, a comprehensive benchmark encompassing four diverse global cities, specifically constructed to assess raw MLLM-driven agents in city navigation. Agents are required to rely solely on visual inputs and internal multimodal reasoning to sequentially navigate $50+$ decision points without additional environmental annotations or specialized architectural modifications. Crucially, agents must autonomously achieve localization through interpreting city-specific cues and recognizing landmarks, perform spatial reasoning, and strategically plan and execute routes to their destinations. Through extensive evaluations, we demonstrate that current state-of-the-art MLLMs, reasoning techniques (e.g., GEPA, chain-of-thought, reflection) and competitive baseline PReP significantly underperform in this challenging setting. To address this, we propose \emph{Verbalization of Path} (VoP), which explicitly grounds the agent’s internal reasoning by probing city-scale cognitive maps (key landmarks and directions toward the destination) from the MLLM, substantially enhancing navigation success.  Project Webpage: \url{https://dwipddalal.github.io/AgentNav/}

\end{abstract}

\section{Introduction}

Pretraining MLLMs on web-scale, interleaved image–text corpora induces broad, transferable world knowledge, enabling robust zero-shot and few-shot generalization across diverse vision–language tasks \cite{alayrac2022flamingo, huang2023language, radford2021learning, dalal2025constructive} thereby facilitating the development of sophisticated agents \cite{xie2024large, hu2025dataset, gao2024multi}. This emergent behavior has enabled foundation models to exhibit robust general reasoning and instruction-following abilities \cite{brown2020language, achiam2023gpt4, touvron2023llama2, yang2024qwen25}. Furthermore, their multimodal successors \cite{openai2024hellogpt4o, reid2024gemini15, liu2024llavanext, wang2024qwen2vl, chen2023internvl}, enhanced by recent scaling methods \cite{chen2025expanding}, integrate high-resolution visual processing, extended context handling, and precise OCR and grounding capabilities. Such advancements facilitate elegant perception-to-action pipelines, enabling researchers to develop embodied agents capable of complex perception, planning, and action tasks \cite{huang2022language, huang2023voxposer, ahn2022can, song2023llm, zhou2024navgpt, lin2025navcot, yue2024mllm}.

\begin{figure*}[!t]
  \centering
  \includegraphics[width=\textwidth]{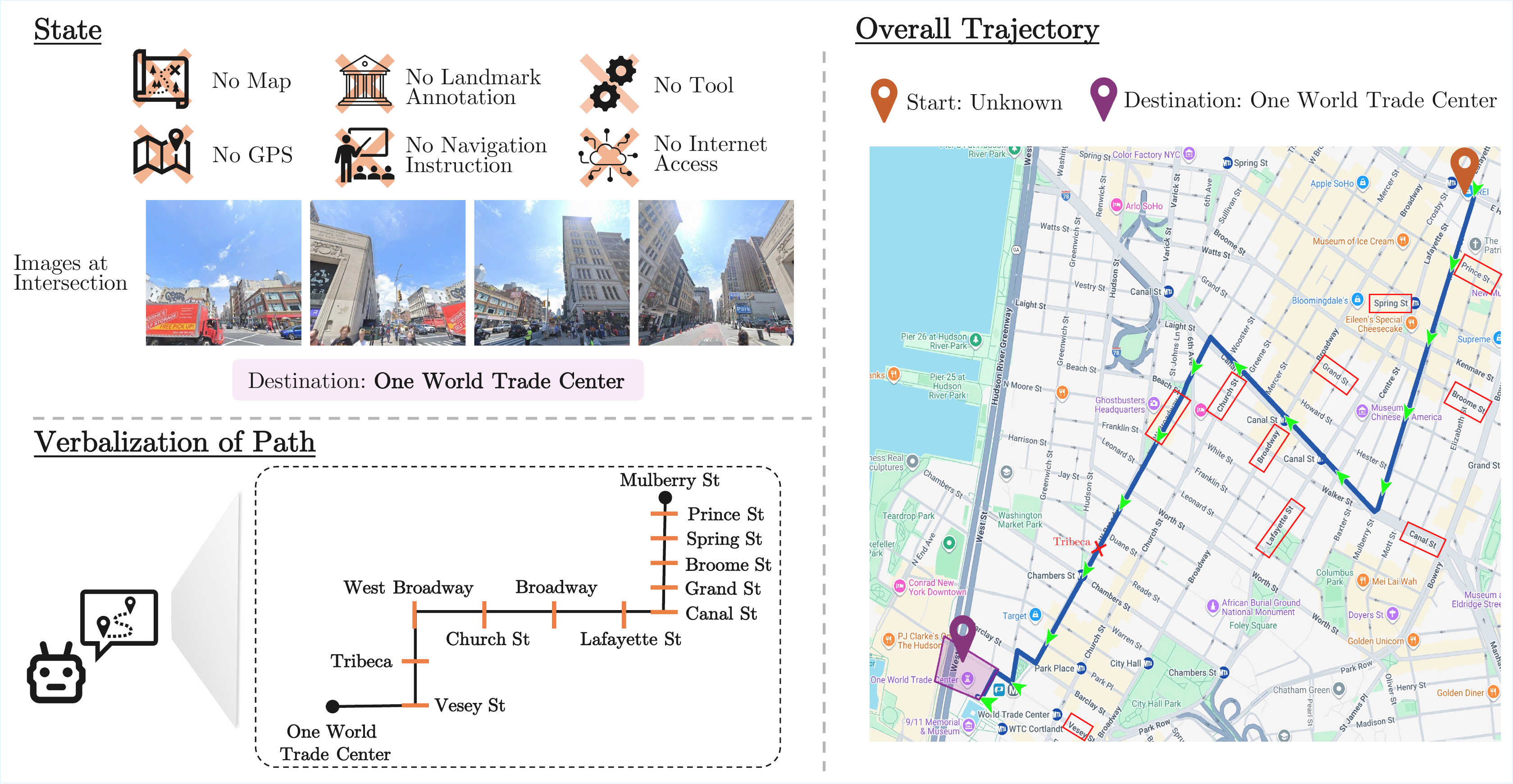}
  \caption{The figure illustrates our proposed \emph{Verbalization of Path} (VoP) method, which elicits city-scale cognitive maps from MLLMs for city navigation in the wild. The red bounding boxes on the New York map highlight the streets and locations explicitly referenced by the MLLM during verbalization of path.}
  \label{fig:vop}
\end{figure*}

While existing works have demonstrated reasoning capabilities of embodied agents in indoor navigation \cite{zhou2024navgpt, lin2025navcot, yue2024mllm}, code generation and planning \cite{song2023llm, liang2022code}, and robotic arm manipulation \cite{huang2023voxposer, ahn2022can, singh2022progprompt}, these evaluations predominantly occur within simulated environments and are not knowledge intensive tasks. Therefore do not rigorously assess the agents' ability to leverage their extensive internal knowledge repositories to execute sequential decisions in dynamic, real-world scenarios.

Whereas, outdoor navigation is inherently a knowledge-intensive task demanding extensive cognitive capabilities, such as comprehensive environmental knowledge, sequential decision-making, spatial reasoning, and robust visual grounding using recognizable landmarks \cite{mirowski2018cities_without_map}. Prior studies on outdoor navigation have typically supplied agents with explicit landmark information embedded within images \cite{chen2019touchdown, zeng2024perceive, schumann2024velma}, significantly alleviating cognitive load by eliminating the necessity for agents to internally retrieve knowledge for self-localization and planning. In contrast, our work introduces a novel task termed \emph{Sparsely Grounded Long-Range Navigation} where:

\begin{taskbox}
Agent must navigate without any landmark annotations or explicit city navigation instructions, relying exclusively on images observed at each intersection. This task requires agents to leverage their intrinsic world knowledge to facilitate spatial understanding, accurate self-positioning, and sequential decision-making to reach the goal.
\end{taskbox}

Existing datasets \cite{mirowski2018cities_without_map, chen2019touchdown} exhibit several limitations: 1) they contain relatively short path lengths, 2) they are restricted to one-two cities, 3) given their widespread usage over the years, there is a significant likelihood that MLLMs have been exposed to these datasets during training. Hence, we introduce a novel, diverse dataset, CityNav, comprising paths of length greater than 2 Km, and include over 50+ decision points, spanning four distinct cities, thereby testing varied capabilities of MLLMs and significantly enhancing the task complexity. Our dataset is constructed using Google Street View panograph \cite{anguelov2010google}. Alongside the dataset, we provide a robust evaluation platform capable of deploying MLLMs directly onto the Google Street View navigation graph. The platform is explicitly designed to handle practical navigation challenges such as dead ends, missing street connections, and abrupt transitions inherent to Google Street View.

We further introduce \emph{Verbalization of Path (VoP)}, a mechanism designed to explicitly extract and leverage the latent world knowledge internalized by MLLMs. By prompting agents to verbalize navigation paths, as illustrated in Fig.~\ref{fig:placeholder}, VoP substantially enhances the performance of MLLM-based agents on long-range navigation tasks. Since navigating unstructured environments is widely regarded as a fundamental hallmark of intelligence~\cite{mirowski2018cities_without_map}, our results highlight the effectiveness of VoP in bridging the gap between static reasoning capabilities and dynamic, real-world sequential decision-making.

\vspace{15pt}
Our main contributions are 1) We introduce a new task and dataset designed to test MLLMs on long-range sequential decision-making that requires leveraging their internal world knowledge. 2) We propose a zero-shot framework \emph{Verbalization of Path} to elicit and utilize the internal world knowledge of MLLMs for effective outdoor navigation. 3) We show that MLLMs can successfully navigate complex urban environments such as New York City, indicating that these models possess extensive structured world knowledge capable of supporting real-world spatial reasoning. 4) We demonstrate that state-of-the-art reasoning techniques (e.g., GEPA, reflection) that are effective for static reasoning fail when MLLMs must elicit and deploy their internal knowledge for dynamic, sequential decision-making.

\begin{table*}[t]
\centering
\footnotesize
\setlength{\tabcolsep}{3pt}
\renewcommand{\arraystretch}{1.05}
\resizebox{0.9\textwidth}{!}{%
\begin{tabular}{l l l l l l l l l}
\toprule
Work & Bucket & World & Scope & Route-length & Visual input & Guidance & Landmark & MLLM role \\
\midrule
EmbodiedBench & Embodied-bench & Sim & Multi-env & Room-scale & RGB & Provided & -- & Raw \\
NavGPT & Indoor VLN & Sim & MP3D & Room-scale & Text-description & Provided & -- & LLM \\
NavCoT & Indoor VLN & Sim & MP3D & Room-scale & RGB & Provided & -- & Fine-tune \\
Touchdown & Outdoor VLN & Real & Manhattan & 350\,m & Panorama &  Provided & Text & -- \\
VELMA & Outdoor VLN & Real & Manhattan & 350\,m & Panorama &  Provided & CLIP & Fine-tune \\
Loc4Plan & Outdoor VLN & Real & M2S & 350\,m & Panorama &  Provided & Locate & Fine-tune \\
VLN-Video & Outdoor VLN & Real & Multi-city & -- & Video &  Provided & Implicit & Fine-tune \\
FLAME & Outdoor VLN & Real & M2S & 350\,m & Panorama &  Provided & Implicit & Fine-tune \\
Perceive-Reflect-Plan & Outdoor VLN & Real & Multi-city & 1.5 km & Scene & Provided & Explicit & Fine-tune \\
\midrule
CityNav (ours) & Outdoor VLN & Real & Multi-city & 2\,km & Intersection-Images & None & None & Raw \\
\bottomrule
\end{tabular}%
}
\vspace{2pt}
\caption{Dataset-level comparison emphasizing how CityNav differs from prior navigation/embodied benchmarks. Abbreviations: MP3D=Matterport3D, M2S = Map2Seq.}
\label{tab:citynav_comparison}
\end{table*}

\section{Related Works}

This section summarizes the closest related works and clarifies how our work differs in setting and design. We discuss embodied environments, MLLMs and sequential decision making, and instruction-based outdoor VLN. The differences are clearly delineated in Table~\ref{tab:citynav_comparison}.

\paragraph{Embodied Environment.}
Existing evaluation suites for embodied agents differ along domain, action granularity, and multimodality. Household high-level instruction-following benchmarks \cite{shridhar2020alfworld,lin2025navcot,zhou2024navgpt,shridhar2020alfred,li2023behavior1k, szot2023llarp,choi2024lotabench,li2024embodiedinterface} focus primarily on symbolic task decomposition and sequencing in indoor scenes, typically within one or two simulators. Finally, multi-domain evaluations at high level split into text-only, LLM-centric AgentBench and multimodal VisualAgentBench, which standardize agent scaffolds but remain at abstract action levels \cite{liu2023agentbench,liu2024visualagentbench}.  \cite{yang2025embodiedbench} unifies these strands with both high- and low-level action spaces and introducing a capability-oriented evaluation protocol. In contrast to simulator-bound or single-skill suites, our \ourmethodshort benchmark places \emph{raw MLLMs} in \emph{real, long-horizon city navigation} directly on the Google Street View graph—\emph{without} landmarks, maps, or auxiliary annotations.

\paragraph{MLLMs and Sequential Decision Making.}

Recent advancements in prompting \cite{wei2022chain,yao2023tree,wang2022self,yao2024got,manas2024cot,wang2023promptagent} have shown LLMs can exhibit sophisticated reasoning, and significantly improve performance on tasks requiring  intermediate reasoning in static environment. \cite{yao2023react, shinn2023reflexion,gao2023pal,agrawal2025gepa,xu2023reprompting} extend this further by iteratively planning through interactive feedback and reflection. In this work, we show that while these methods perform effectively in static contexts, they degrade significantly in sequential decision-making tasks that require methods to \emph{coax} the internal world knowledge.  

\paragraph{Instruction-based outdoor navigation.}
Vision-and-Language Navigation (VLN) \cite{gu2022vision} addresses the challenge of jointly grounding linguistic instructions and visual perception in realistic environments. Prior works \cite{chen2019touchdown,schumann2021map2seq,schumann2022outdoor_generalization} introduced landmark-rich navigation dataset. \cite{hermann2020learning} aligned textual instructions with visual observations in partially observable Street View environments, while \cite{mirowski2018learning} employed reinforcement learning to improve navigation robustness. More recent methods \cite{schumann2024velma, zeng2024perceive, xu2025flame, tian2024loc4plan} fine-tune MLLMs with city landmark–based instruction following, whereas \cite{li2024vlnvideo} leverages driving videos to provide dense visual supervision for route following. All prior approaches rely on explicit textual or landmark-based instructions and operate over short trajectories (typically under 350 m). In contrast, our dataset \ourmethodshort focuses on long-range navigation (average path length $\approx$ 2 km) and provides no auxiliary environmental information—only images at each intersection—requiring the model to infer spatial relations, coax its internal world knowledge and plan trajectories.

\section{\ourmethodshort}

We model autonomous city navigation as a Partially Observable Markov Decision Process (POMDP) defined on an undirected graph \(G = (V, E)\), where \(V\) denotes intersections and \(E\) represents undirected street segments, each associated with a positive length \(\ell(e) > 0\) for \(e \in E\).

At any given discrete time step \(t\), the state \(s_t\) of the agent corresponds directly to its current intersection: $s_t = v_t \in V.$ From an intersection \(v_t\), the set of available actions is defined as the set of street segments incident to \(v_t\): $\mathcal{A}(v_t) = \{ e \in E \mid v_t \in e \}$. When an agent at intersection \(v_t\) selects an action \(a_t = e \in \mathcal{A}(v_t)\), it deterministically transitions to the adjacent intersection \(v'\) connected by the chosen street segment \(e = \{v_t, v'\}\). 

The system exhibits \emph{sparse grounding}, as visual observations are only available at intersections and entire state in form of map is not available. Specifically, upon reaching an intersection \(v_t\), the agent receives a set of images: $o_t = \{ I_e \mid e \in \mathcal{A}(v_t) \},$ where each image \(I_e\) corresponds to visual input associated with the street segment represented by edge \(e\). Between intersections, while navigating along a street segment, the agent receives no visual observations.

Action selection at each intersection is governed by a policy \(\pi\), defined as a mapping from the current state \(v_t\) and observation set \(o_t\) to an action \(a_t\): $\pi: V \times O \rightarrow E, \quad a_t = \pi(v_t, o_t),$ where \(O\) represents the space of possible observation sets. The policy thus dictates the agent’s decision-making process, leveraging available visual information to select the next street segment to traverse.

\subsection{Dataset}

We curated a diverse dataset explicitly designed to evaluate multiple dimensions of navigation planning in MLLM-based embodied agents. Our city selection methodology targeted locations exhibiting considerable diversity across primary language usage, architectural style, signage characteristics, and street layout topology, including variations in grid density and road complexity. This strategic diversity ensures exposure of the model to a comprehensive array of real-world navigational scenarios.

Specifically, we selected four globally distributed cities, each presenting distinct navigational challenges designed to rigorously assess the adaptability of the model. For example, Tokyo, Japan, predominantly employs Japanese-language signage and place nomenclature, thus posing significant linguistic barriers to LLMs primarily trained on English-dominated corpora. Within each city, we systematically identified and annotated 100 distinct origin-destination pairs, forming standardized evaluation tasks.

\textbf{Manual Annotation.} The destinations are not always just a building, sometimes they are as big as park. So in those cases, the destination doesn't necessarily need to be just a node, it can be collection of nodes. Since it's not possible to figure this out algorithmically, we manually annotate the destinations for each of the chosen places. So we draw a polygon around the destination that acts as a boundary for the destination and when agent reaches this boundary we call it reached.

\begin{figure}[t]
    \centering
    \includegraphics[width=\linewidth]{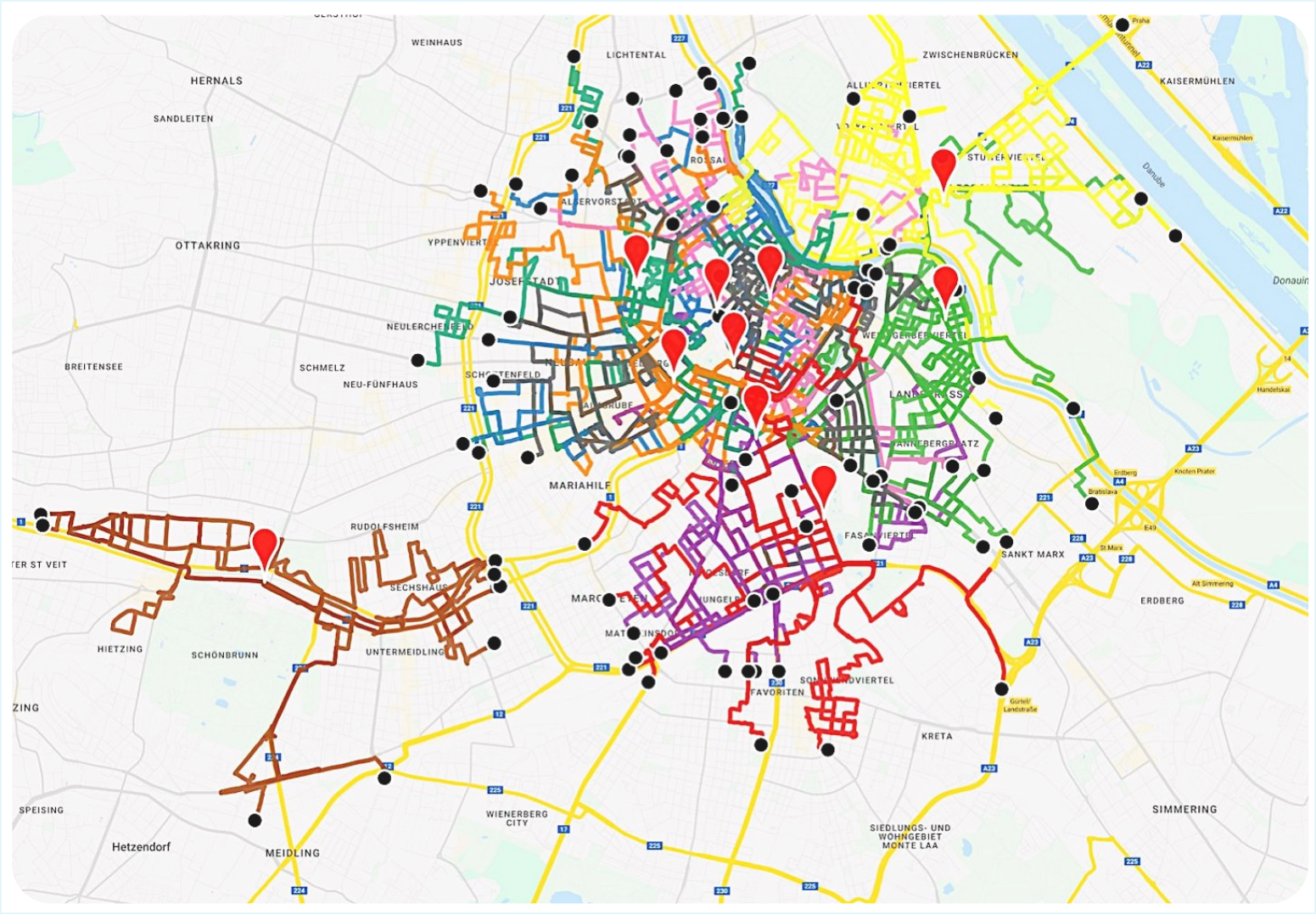}
    \caption{Dataset paths visualization of Vienna. Here the black dots mark the starting point, and the red blobs mark the destination point.}
    \label{fig:placeholder}
\end{figure}

\textbf{Random Sampling of Starting Location.} Starting from a seed node $v_s \in V$, our crawler aims to reach a target radial distance $d_{\text{target}}$ from $v_s$. The traversal operates in two distinct phases: first, a deterministic corridor-following phase continues along nodes having an effective out-degree of $1$ (excluding the backward link), until encountering the first decision junction (nodes with out-degree $\geq 2$). The second phase involves a depth-first search (DFS) using an explicit junction stack with backtracking. At each junction node $v_j$, the crawler chooses among candidate edges $e_j^i \in E(v_j)$ according to a probability distribution computed via softmax over their angular deviation $\theta_i$ relative to the desired heading (typically directed away from the seed node). Specifically, the selection probability is: $P(e_j^i) = \frac{\exp(\cos(\theta_i)/T)}{\sum_k \exp(\cos(\theta_k)/T)}$ where the temperature parameter $T$ anneals with increasing straight-line distance $d(v_j, v_s)$ from an initial random exploration ($T \to \infty$) toward a progressively more directional selection (lower $T$). To mitigate loops and encourage diverse coverage, we impose an exponential revisit penalty factor $\gamma^{n_v}$ to the selection probability, where $n_v$ is the visit count for node $v$, and $0 < \gamma < 1$. Once the crawler reaches the radial target distance $d_{\text{target}}$, it optionally continues for a small number of steps to terminate at a node with degree greater than or equal to a threshold $d_{\text{min\_final}}$.

\textbf{Google Street View Pre-processing.} 
We enhance the underlying Google Street View graph for reliable navigation tasks, we systematically identify and resolve common structural issues. One of the primary challenges in constructing our navigation graph from Google Street View data arises from structural inconsistencies such as dead ends, incomplete coverage, and asymmetric links. We define a \emph{dead end} as a node with an outgoing edge to a neighboring node that does not lead to any further intersections (i.e., a terminal street segment with no valid successors). To ensure graph connectivity and eliminate such artifacts, we algorithmically identify and prune dead ends during preprocessing. Additionally, the Street View panograph often exhibits asymmetric connectivity, where a link from node \(a\) to node \(b\) exists, but the reverse link from \(b\) to \(a\) is absent. This breaks the undirected graph assumption required for consistent navigation. To resolve this, we crawl the underlying graph and explicitly add the missing reverse edges whenever such inconsistencies are detected, thereby restoring bidirectional connectivity and ensuring that the resulting graph is well-formed for navigation tasks.

\begin{table}[t!]
\centering
\resizebox{\columnwidth}{!}{%
\begin{tabular}{l c p{0.6\linewidth} c c c}
\toprule
\textbf{Cities} & \textbf{Region} & \textbf{Diversity} & \textbf{Distance} & \textbf{Decision Points} \\
\midrule
New York   & USA     & Grid-Based, Well Spaced, Rich Street Signs & 1.8  & 44  \\
São Paulo  & Brazil  & Non-Block Structure, Portuguese Language                         & 2.0   & 55  \\
Tokyo      & Japan   & Short Sightlines in Narrow Alleys, Japanese Language      & 1.9 & 80  \\
Vienna     & Austria & Road blocks because of rails, German Language                   & 2.1 &  60 \\
\bottomrule
\end{tabular}%
}
\caption{Dataset statistics across four cities. \emph{Diversity}: qualitative descriptors of urban form and visual/linguistic variety that affect navigation (e.g., grid regularity, sightlines, signage language). \emph{Distance}: average path length (km) for routes in our test split. \emph{Decision Points}: mean number of discrete navigation decisions per route (intersections).}

\label{tab:city_metrics}
\end{table}

\section{\ourmethodagent}

\begin{figure*}
    \centering
    \includegraphics[width=\linewidth]{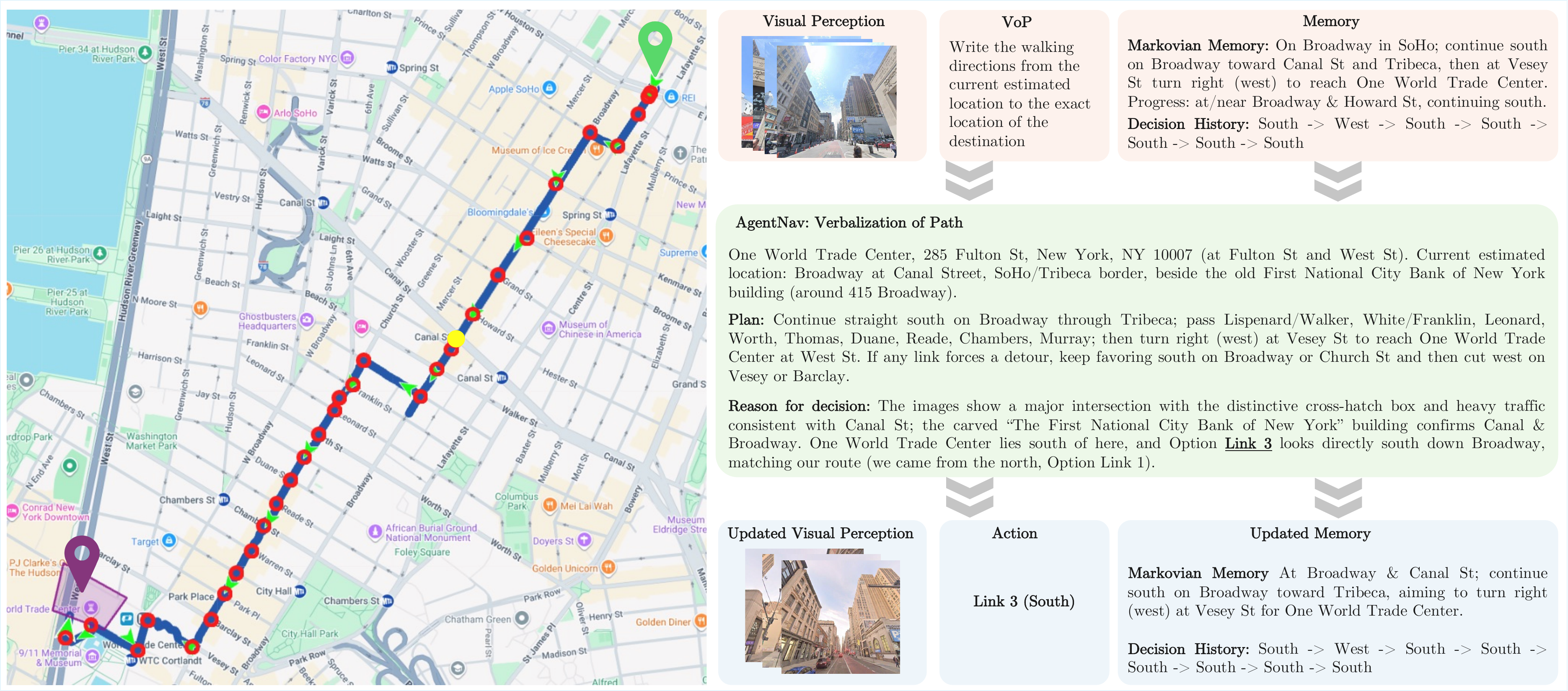}
    \caption{Illustration of state transition from \( S \) (marked by the yellow dot) to \( S+1 \) using the agent’s internal reasoning through \emph{Verbalization of Path}. At each state, the agent perceives visual cues and references its memory to update decisions and navigation strategy. The purple marker denotes the destination (One World Trade Center), while the green marker indicates the starting point.}
    \label{fig:placeholder}
\end{figure*}

\subsection{Grounding with Verbalized Paths}

Successful outdoor navigation fundamentally requires accurate self-localization and comprehensive world knowledge. Here, we demonstrate a targeted approach to probe such knowledge explicitly from MLLMs. We augment the agent's prompt-as-policy framework with three distinct phrases that consistently elicit robust navigation performance by explicitly grounding the agent's internal state and reasoning in the external world. Specifically, we incorporate the following structured prompts: 1) \emph{Write the exact location of the destination}: This explicitly defines the navigation goal, anchoring the agent's decision-making process to a clear terminal state.
2) \emph{Write the current estimated exact location}: This compels the agent to continuously estimate and update its current position, serving as a precise initial condition for subsequent decisions.
3) \emph{Write the walking directions from the current position to the destination}: Crucially, this leverages the agent's generalist knowledge, prompting it to generate actionable instructions grounded in real-world spatial relationships and pathfinding logic.

\subsection{Memory of \ourmethodagent}
For extensive runs averaging over 50+ decisions per trajectory, maintaining an efficient memory system is critically important. Traditional episodic memory architectures which store comprehensive information (images, decisions, analyses) for each step across multiple past episodes quickly become computationally intractable, scaling exponentially with episode length. To mitigate this issue, we strategically decompose memory management into three core components that significantly enhance efficiency (approximately a 100-fold reduction in memory overhead) within a Partially Observable Markov Decision Process (POMDP) framework: \textit{Markovian Memory}, \textit{Decision History}, and \textit{Previous Visit Tracking}.

\paragraph{Markovian Memory.}

We implement \textit{Markovian Memory} by explicitly prompting the agent to produce a memory state at each decision step. Formally, at time step \(t\), the agent's input includes the previous memory state \(m_{t-1}\), and along with next action \(a_t\) it outputs updated memory state \(m_t\). This process can be expressed as: $(a_t, m_t) = \pi(v_t, o_t, m_{t-1})$ where \(m_t\) represents a sufficient statistic summarizing past observations, effectively transforming the partially observable process into a Markovian one within an augmented state space \(\tilde{s}_t = (v_t, m_t)\). This mechanism eliminates the need for full episodic memory, thereby significantly reducing computational and storage costs. Furthermore, as the model becomes increasingly capable, it learns to selectively preserve only the most relevant information for effective decision-making, resulting in a more compact and adaptive memory representation.

\begin{table*}[t]
\centering
\resizebox{\textwidth}{!}{%
\setlength{\tabcolsep}{6pt}
\begin{tabular}{ll ccc ccc ccc ccc}
\toprule
 & & \multicolumn{3}{c}{New York} & \multicolumn{3}{c}{Tokyo} & \multicolumn{3}{c}{Vienna} & \multicolumn{3}{c}{Sao Paulo} \\
\cmidrule(lr){3-5}\cmidrule(lr){6-8}\cmidrule(lr){9-11}\cmidrule(lr){12-14}
\textbf{MLLM} & \textbf{Agent Config} &
\textbf{Success} & \textbf{SPL} & \textbf{D.A.} &
\textbf{Success} & \textbf{SPL} & \textbf{D.A.} &
\textbf{Success} & \textbf{SPL} & \textbf{D.A.} &
\textbf{Success} & \textbf{SPL} & \textbf{D.A.} \\
\midrule
\multirow{2}{*}{GPT 4o} & Base & 13 & 0.064  & 39.04  & 4 & 0.046 & 36.79 & 4 & 0.031 & 35.67 & 3 & 0.040 & 34.69 \\
 & AgentNav & 88 & 0.539 & 72.91 & 14 & 0.099 & 40.90 & 26 & 0.170 & 46.32 & 20 & 0.06 & 43.45 \\
\midrule
\multirow{2}{*}{GPT 5} & Base & 54  & 0.375  & 55.97 & 10 & 0.088 & 41.24 & 11 & 0.092 & 40.72 & 7 & 0.051 & 36.98 \\
 & AgentNav & 94 & 0.711 & 82.98 & 30 & 0.163 & 54.97 & 56 & 0.226 & 54.82 & 29 & 0.126 & 48.96 \\
\midrule
\multirow{2}{*}{GPT 4.1} & Base & 15 & 0.097 & 42.27 & 5 & 0.044 & 38.83 & 2 & 0.037 & 34.66 & 5 & 0.049 & 35.46 \\
 & AgentNav & 92 & 0.557 & 75.27 & 17 & 0.101 & 43.67 & 32 & 0.182 & 49.95 & 22 & 0.080 & 44.05 \\
\midrule
\multirow{2}{*}{O3} & Base & 48 & 0.490 & 64.36 & 7 & 0.049 & 40.33 & 9 & 0.083 & 39.82 & 6 & 0.075 & 35.68 \\
 & AgentNav & 95 & 0.759 & 84.93 & 27 & 0.142 & 52.56 & 38 & 0.190 & 50.73 & 24 & 0.117 & 50.75 \\
\midrule
\multirow{2}{*}{Gemini 2.5 Flash} & Base & 12 & 0.060 & 41.57 & 8 & 0.049 & 39.97 & 1 & 0.010 & 29.31 & 5 & 0.049 & 35.80 \\
 & AgentNav & 73 & 0.471 & 74.75 & 17 & 0.066 & 46.87 & 17 & 0.137 & 46.35 & 12 & 0.085 & 43.65 \\
\midrule
\multirow{2}{*}{Qwen 2.5 VL 32b} & Base & 7 & 0.089 & 35.11 & 2 & 0.023 & 30.01 & 0 & 0.0 & 26.1 & 2 & 0.011 & 29.87 \\
 & AgentNav & 32 & 0.153 & 56.39 & 12 & 0.094 & 40.03 & 12 & 0.119 & 44.94 & 9 & 0.059 & 37.80 \\
\bottomrule
\end{tabular}
}
\caption{Base model vs.\ \ourmethodagent across four cities. We report Success, SPL(Success weighted by Path Length), and D.A. (Decision Accuracy); higher is better. \ourmethodagent consistently and substantially improves performance over the base MLLM across all model families, indicating importance of \ourmethod method.}
\label{terimakins}
\end{table*}

\paragraph{Decision History.} It maintains a structured record of the sequence of actions chosen by the agent at each intersection during the trajectory. Formally, this can be represented as:
$\mathcal{H}_t = \{a_1, a_2, \dots, a_t\}$. Maintaining this ordered sequence enables the agent to leverage its own behavioral trajectory for reasoning about prior choices, route corrections, and avoid repeated loops. By focusing on compact action traces instead of complete episodic histories, this mechanism provides a balance between computational efficiency, and long-horizon reasoning.

\paragraph{Previous Visit.}

It provides the agent with awareness of past interactions at specific intersections. Each time the agent revisits an intersection \(v_t\), it retrieves the record of its previous decisions taken at that node, enabling it to reason about prior outcomes. Repeatedly encountering the same intersection typically indicates that the agent is caught in a local loop or has failed to make progress toward the destination. To mitigate this, the system encodes a visit count \(n_v\) for each node \(v\), which influences the policy’s exploration behavior. As \(n_v\) increases, the agent is progressively discouraged from repeating the same action-promoting exploration and preventing cyclic behavior. For example, if the agent has chosen to go west multiple times from an intersection without improvement, the memory mechanism biases future actions toward unexplored directions such as east. This structured representation of visit history thus endows the agent with self-awareness of its traversal patterns, improving navigational robustness in complex city graphs.

\section{Experimentation}
In this section, we present a comprehensive empirical evaluation of AgentNav. We benchmark six strong MLLMs and multiple state-of-the-art reasoning and navigation baselines on CityNav. We then quantify the contribution of each component through ablation studies (Sec.~\ref{ablate}). We provide an error analysis and qualitative failure cases to characterize the remaining limitations (Sec.~\ref{error}).

\subsection{Implementation Details}

For each run within every city, we set a maximum limit of 150 decision points for the agent to reach its destination before automatic termination. Additionally, we restrict the maximum number of graph node transitions to 2000. Rather than executing self-positioning at every decision point, we perform self-positioning every third decision point (a separate call to same MLLM). This deliberate choice introduces minor positional uncertainty, effectively testing the robustness and accuracy of the verbalized path by challenging the agent to reason with slightly imprecise localization.

\subsection{MLLMs and Reasoning Baselines}

We evaluate our method using a selection of strong closed-source and open-source multimodal foundation models to effectively probe their internal world knowledge and reasoning capabilities. Specifically, our evaluation includes: GPT-4o \cite{achiam2023gpt}, a widely-used baseline model in multimodal research; GPT-4.1, which is expected to demonstrate enhanced geographical reasoning capabilities~\cite{grainge2025assessing}; Gemini-2.5 Flash \cite{team2024gemini}, serving as an additional closed-source comparison; GPT-5 (thinking) \& O3, known for its advanced reasoning abilities; and Qwen-2.5VL-32B \cite{bai2025qwen2}, a powerful open-source counterpart. Our focus on these sophisticated multimodal models is motivated by our goal of \emph{coaxing} out the latent world knowledge embedded within large-scale, web-trained models.

To systematically evaluate reasoning effectiveness, we benchmark our approach against state-of-the-art reasoning baselines, including GEPA~\cite{agrawal2025gepa}, Chain-of-Thought (CoT)~\cite{wei2022chain}, Self-Reflection (GPT-4.1)~\cite{shinn2023reflexion}, and Self-Reflection (GPT-5). Here, the labels GPT-4.1 and GPT-5 denote the specific models employed during the reflective reasoning step, wherein the initial reasoning output is revisited and refined. Additionally, we compare our method with the state-of-the-art outdoor navigation baseline, PReP~\cite{zeng2024perceive}. 

\begin{table*}[ht]
\centering
\resizebox{\textwidth}{!}{%
\setlength{\tabcolsep}{6pt}
\begin{tabular}{l ccc ccc ccc ccc}
\toprule
 & \multicolumn{3}{c}{New York} & \multicolumn{3}{c}{Tokyo} & \multicolumn{3}{c}{Vienna} & \multicolumn{3}{c}{Sao Paulo} \\
\cmidrule(lr){2-4}\cmidrule(lr){5-7}\cmidrule(lr){8-10}\cmidrule(lr){11-13}
\textbf{Method} & \textbf{Success(\%)} & \textbf{SPL} & \textbf{D.A.(\%)}
& \textbf{Success(\%)} & \textbf{SPL} & \textbf{D.A.(\%)}
& \textbf{Success(\%)} & \textbf{SPL} & \textbf{D.A.(\%)}
& \textbf{Success(\%)} & \textbf{SPL} & \textbf{D.A.(\%)} \\
\midrule
GPT-4.1 & 15 & 0.097 & 42.27 & 5 & 0.044 & 38.83 & 2 & 0.037 & 34.66 & 5 & 0.049 & 35.46 \\
CoT  & 21 & 0.173 & 44.59 & 9 & 0.077 & 41.09 & 4 & 0.039 & 34.88 & 7 & 0.055 & 37.93 \\
Self Reflection (GPT-4.1) & 16 & 0.112 & 42.90 & 4 & 0.040 & 36.20 & 3 & 0.042 & 36.33 & 12 & 0.052 & 41.95 \\
Self Reflection (GPT-5) &  22 & 0.168 & 48.14 & 8 &  0.079 & 41.48 & 5 & 0.045 & 37.84 & 13 & 0.050 & 41.64 \\
GEPA & 37 & 0.251 & 43.24 & 10 & 0.036 & 42.97 & 5 & 0.013 & 39.74 & 17 & 0.093 & 40.21 \\
PReP & 39 & 0.248 & 36.07 & 5 & 0.010 & 40.68 & 5 & 0.025 & 38.11 & 22 & 0.157 & 41.11 \\
\textbf{\ourmethodagent} & 92 & 0.557  & 75.27  & 17 & 0.101 & 43.67 & 32 & 0.182 & 49.95   & 22 & 0.080 & 44.05 \\
\bottomrule
\end{tabular}%
    }
\caption{The table shows comparison results with different baselines. All the experiments here are performed using GPT-4.1. Self Reflection (GPT-4.1) means the agent used is GPT-4.1 and reflection is done with GPT-4.1. Self Reflection (GPT-5) agent used is GPT-4.1 and reflection is done with GPT-5.}
\label{mishra}
\end{table*}

\subsection{Evaluation Metrics}
We employ standard reasoning and navigation metrics:

\noindent$\bullet$~\textbf{Success:} If the agent reaches the destination node successfully, it receives a score of 1; otherwise, the score is 0.
    
\noindent$\bullet$~\textbf{SPL (Success weighted by Path Length):} SPL evaluates the agent's navigation efficiency by comparing the optimal (shortest possible) path distance \( d_{\text{opt}} \) to the actual distance traveled by the agent \( d_{\text{agent}} \), scaled by the binary success indicator \( S \in \{0,1\} \). $\text{SPL} = S \times \frac{d_{\text{opt}}}{\max(d_{\text{agent}}, d_{\text{opt}})}$ where \( S = 1 \) if the agent successfully reaches the destination, and \( S = 0 \) otherwise.

\noindent$\bullet$~\textbf{Decision Accuracy (D.A.):} The percentage of correct navigation decisions (e.g., correct turns at junctions) made by the agent. A decision is classified as correct if the remaining walking distance to the destination (calculated using the Google Street View API) decreases after executing that decision.

\subsection{Quantitative Results}

\paragraph{Significant Performance Across 5 MLLMs.}
Table~\ref{terimakins} demonstrates that VoP consistently delivers strong improvements across all five evaluated MLLMs. In high-density urban settings such as New York, our approach achieves significantly higher success rates and SPL scores compared to the base configurations of each model. This indicates that the \emph{Verbalization of Path} mechanism effectively coaxes latent world knowledge from MLLMs, enabling more reliable decision-making in complex navigation tasks. Importantly, similar performance gains are observed across all four cities, highlighting the generality of VoP.

\paragraph{Limitations of Existing Reasoning Techniques and Navigation baseline.}
Table \ref{mishra} highlights the limitations of state-of-the-art reasoning strategies. While these techniques have proven effective for static reasoning tasks, they fail to achieve competitive performance on our long-range, embodied navigation benchmark. Their inability to elicit sufficiently structured internal knowledge from MLLMs leads to lower success rates.

\paragraph{Challenges Arising from the Diversity of \ourmethodshort.}
Despite the substantial gains delivered by \ourmethodagent, Table ~\ref{terimakins} also reveals the inherent difficulty of long-horizon sequential decision-making tasks. Even with \emph{Verbalization of Path}, absolute success rates and SPL scores remain modest for Tokyo, Vienna and Sao Paulo, reflecting the persistent gap in the reasoning abilities of current MLLMs and reasoning methods (see Table \ref{mishra}).

\subsection{Ablations and Analysis}
\label{ablate}

We systematically analyze the contributions of individual components of our method, as detailed in Table~\ref{tab:ablation}. Starting with the GPT-4.1 baseline, we observe a success rate of \(15\%\). Incrementally integrating memory components highlights their individual impacts: incorporating \textit{Markovian Memory} increases success to \(23\%\), and further addition of \textit{Decision History} boosts performance to \(29\%\). Finally, integrating \textit{Previous Visit} yields an additional enhancement, reaching a success rate of \(35\%\).

\begin{table}[t]
\centering
\resizebox{\columnwidth}{!}{%
\begin{tabular}{lccc}
\toprule
\textbf{Methods} & \textbf{Success(\%)} & \textbf{SPL} & \textbf{D.A.(\%)} \\
\midrule
GPT-4.1 & 15 & 0.097 & 42.27 \\
GPT-4.1 + Markovian Memory & 23 & 0.162 & 47.19 \\
GPT-4.1 + Decision History & 29 & 0.228 & 55.63 \\
GPT-4.1 + Previous Visit & 35 & 0.298 & 56.67 \\
GPT-4.1 + Partial VoP & 66 & 0.469 & 63.48 \\
\textbf{\ourmethodagent} & \textbf{92} & \textbf{0.557} & \textbf{75.27} \\
\bottomrule
\end{tabular}%
}
\caption{Performance breakdown showing contributions of different components of the \ourmethodagent.}
\label{tab:ablation}
\vspace{-10pt}
\end{table}

Next, we further dissect the VoP method by breaking it down into two stages (Partial VoP and VoP) to clearly identify its effect. In the \textit{Partial VoP} scenario, we instruct the agent to explicitly ground its reasoning solely based on the final destination, prompting it to write the destination's address at the beginning of its reasoning. This targeted grounding notably enhances the success rate to \(66\%\). Subsequently, employing the complete verbalization, where the agent explicitly generates detailed walking directions from its current location to the target destination, our \textit{AgentNav} model reaches the highest performance with a \(92\%\) success rate. This stepwise analysis underscores how each introduced stage of verbalization progressively grounds and elicits richer world knowledge from MLLMs, significantly improving the agent’s capability to reason and navigate reliably in complex, real-world scenarios.

\subsection{Error Analysis \& Failure Cases}
\label{error}

\begin{figure}
    \centering
    \includegraphics[width=0.9\linewidth]{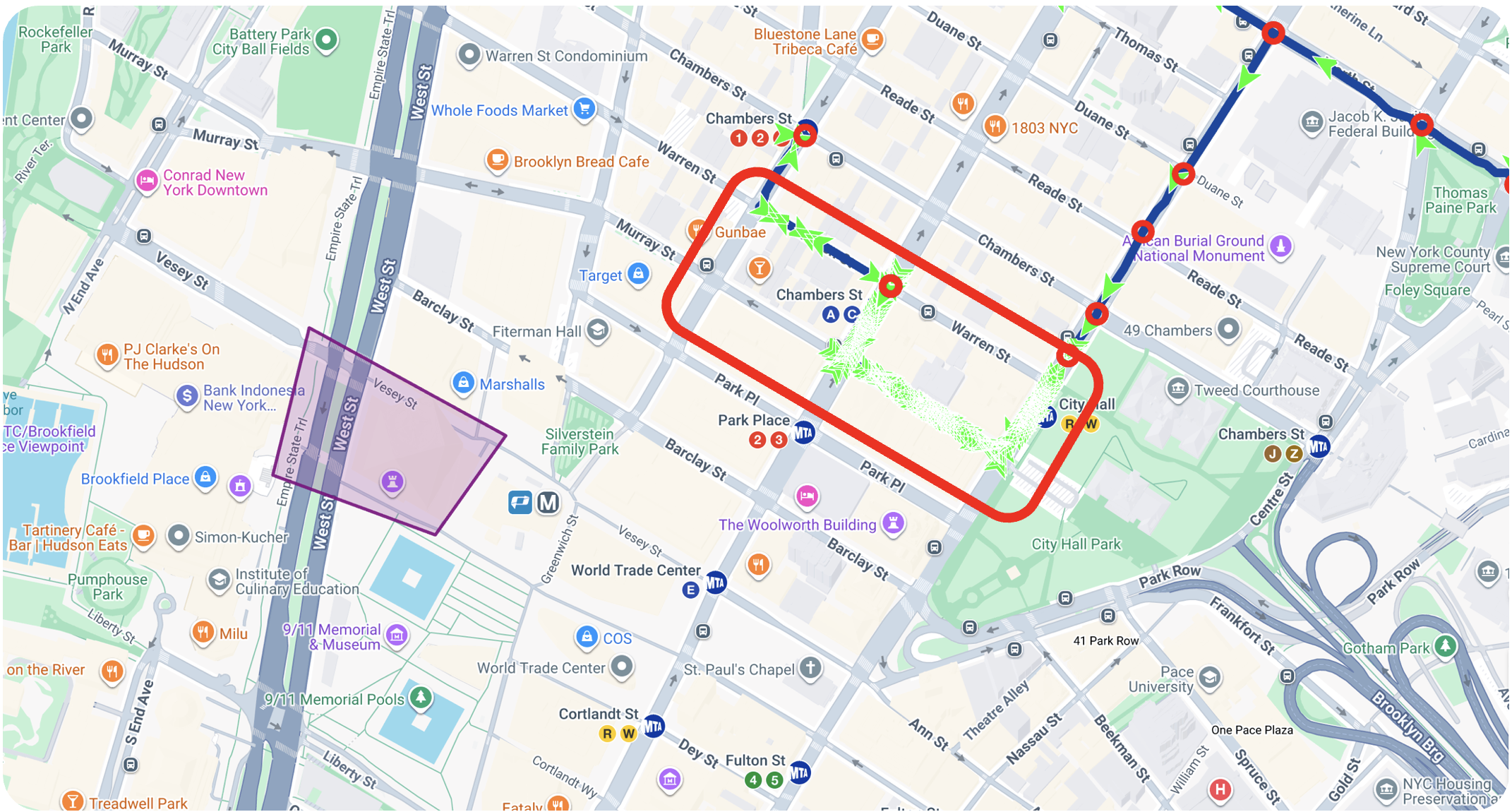}
    \caption{Failure case, the red box highlights the path in which \ourmethodagent gets stuck in a loop.}
    \label{fig:error_figure}
\end{figure}

\paragraph{Grounding Challenges in Complex Linguistic Environments.}
 Occasional failure mode occurs in cities featuring complex linguistic environments, such as Tokyo. Here, visual signage predominantly comprises intricate Kanji characters, which hinders effective visual grounding and accurate self-positioning; causing it to become stuck in a loop (see Figure~\ref{fig:error_figure}).

\paragraph{Over-Reliance on Initial Plans.}
A further failure mode emerges when the agent occasionally demonstrates excessive adherence to its initial navigation plan. In such cases, the agent sometimes prioritizes its original intended route without adequately reconsidering or revising it in response to new observations or emerging evidence.

\section{Discussion \& Future Works}
We highlight two experimental phenomena that point to promising future directions.

\paragraph{(1) Reflection baselines degrade in long-horizon, real-world navigation.}
Reflection-based baselines (e.g., GEPA and reflection-style prompting) work well in static or short-horizon settings but substantially underperform on our real-world, long-horizon benchmark: GEPA achieves only \textbf{37\%} success, and Self-Reflection achieves only \textbf{22\%} success even when the reflection step uses a stronger model (e.g., \textbf{GPT-5}). The key takeaway is that \emph{self-reflection-based approaches do not yield good performance} when success depends on sustaining correctness over many sequential decisions. Overall, building self-evolving agents that genuinely learn from their own rollouts remains a major open challenge in real-world, long-horizon settings.

\paragraph{(2) Task decomposition becomes crucial under long context and long horizons.}
Long trajectories quickly fill the prompt with episodic details, making it harder for the model to preserve the most relevant global state (where am I, what is the plan, what progress have I made?). We find that \emph{distributing the workload} improves performance: we decouple self-positioning from action selection by making a separate model call for localization (performed periodically), then use the resulting estimate to guide route planning and step-level decisions. This matches a broader observation in multi-agent systems: dividing a complex task into subproblems can improve robustness \vs a single monolithic agent. Cognitively, this resembles \emph{executive functions} that coordinate and allocate resources across subtasks \cite{deLangis2025strong}. Future work could study more principled decompositions and how modularization interacts with memory compression and uncertainty estimation.

\section{Conclusion}

In this work, we introduced CityNav, a comprehensive benchmark designed to rigorously assess MLLMs on real-world, long-range urban navigation tasks. Through extensive experiments across diverse global cities, we demonstrated that our proposed \emph{Verbalization of Path} mechanism, complemented by strategic memory components, effectively coaxes the intrinsic world knowledge from MLLMs, resulting in significant performance improvements over existing reasoning methods. These findings underline current limitations of MLLMs for embodied sequential decision-making tasks, emphasizing the need for continued research in robust, adaptive reasoning frameworks.

\section{Limitations}

While CityNav and our proposed AgentNav framework represent a significant step forward in evaluating MLLMs on real-world long-range navigation tasks, several limitations remain. First, despite explicit verbalization and memory mechanisms, absolute success rates and SPL scores remain modest for certain models and cities, highlighting the persistent gap between current MLLM capabilities and the demands of large-scale, embodied reasoning tasks. Second, our evaluation relies on Google Street View imagery and its underlying graph structure which occasionally contain structural inconsistencies such as missing reverse links or misleading panorama orientations. Although our preprocessing mitigates many of these issues, residual anomalies may still introduce unintended noise into the agent’s input and affect its performance. Finally, our benchmark emphasizes four diverse cities but is not yet exhaustive of all possible urban environments, languages, or signage complexities. Models trained primarily on English-dominated corpora may still exhibit limited generalization to regions with highly complex scripts or low-visibility environments.

\bibliography{custom}

\clearpage
\setcounter{page}{1}
\onecolumn
\appendix

\setcounter{section}{0}   
\renewcommand{\thesection}{\arabic{section}}  %
\section*{Appendix}

\begin{itemize}
  \item \textbf{1. Google Street View Policy Compliance Statement}
  \item \textbf{2. Additional Details}
  \item \textbf{3. Improvements to Google StreetView Base Graph}
  \item \textbf{4. Strong Prompt-based Baselines}
  \item \textbf{5. AgentNav Prompt}
  \item \textbf{6. Baseline Prompts}
  \item \textbf{7. Examples}
\end{itemize}

\section{Google Street View Policy Compliance Statement}
Our study uses Google Street View imagery only at evaluation time and only via official Google‑provided interfaces. The evaluation platform executes the agent directly on the Street View navigation graph and fetches the panorama views transiently for decision making; we do not scrape, bulk‑download, mirror, or redistribute Street View imagery. The public release accompanying this paper consists of annotations (origin–destination pairs, destination polygons), code, and evaluation scripts. No images will be released. Human annotation was performed by the authors and is limited to drawing destination boundaries; no personally identifying information was collected or added. As documented in the paper (dataset description in Section 3.1; summary statistics in Table 1), imagery appears only as on‑the‑fly observations at intersections, multilingual scene text may be visible within those images, and all faces and license plates remain blurred as provided by Google, since we do not alter or post‑process Street View content.

To align fully with Google’s Street View and Maps Platform Terms of Service, we (i) access imagery solely through authorized APIs/viewers; (ii) preserve all provider attribution, blurring, and watermarks returned by the service; (iii) avoid storing or caching raw imagery beyond ephemeral runtime needs; (iv) release no derivative image dataset (only text/graph metadata and author‑created annotations under our license); (v) require downstream users to supply their own API keys and to accept and comply with Google’s Terms when reproducing our results; and (vi) forbid any use of our code or annotations to scrape, de‑watermark, reverse‑engineer, or otherwise circumvent Google’s technical and policy safeguards. Our figures, where illustrative thumbnails are necessary for scholarly reporting, are minimal and strictly for explanation of results; they do not constitute redistributable imagery or a dataset. These measures, together with our annotations‑only release and authors‑only human labeling protocol, ensure that the work adheres to Google’s Street View policy while enabling reproducible research on long‑range, real‑world navigation.

\section{Additional Details}

\paragraph{LLM parameters.}
\noindent\textbf{OpenAI (GPT-4o, GPT-4.1; non-reasoning):}
\texttt{temperature=1.0}, \texttt{top\_p=1.0}, \texttt{presence\_penalty=0}, \texttt{frequency\_penalty=0}, \texttt{n=1}, \texttt{stream=false}, \texttt{max\_tokens=8000}. \\
\textbf{Gemini 2.5 Flash:}
\texttt{temperature=1.0} (0--2), \texttt{top\_p=0.95} (0--1), \texttt{top\_k=64}, \texttt{candidate\_count=1}, \texttt{max\_tokens=8000}. \\
\textbf{Qwen 2.5-VL (FP16; Ollama):}
\texttt{temperature=0.8}, \texttt{top\_p=0.9}, \texttt{top\_k=40}, \texttt{repeat\_penalty=1.1}, \texttt{repeat\_last\_n=64}, \texttt{num\_predict=-1}.

\paragraph{Streetview API Parameters.}
The Street View Static API parameters were as follows:
\begin{itemize}
  \item \texttt{size=512x512} (pixels): output resolution of each crop.
  \item \texttt{fov=90} (degrees, horizontal field of view): the angular width of the panorama that is projected into the image (i.e., how wide the virtual camera ``sees'').
  \item \texttt{pitch=+30} (degrees): camera tilt relative to the horizon; positive values tilt upward.
\end{itemize}

The API returns a rectilinear view extracted from the underlying panorama. We set
\texttt{fov=90}$^\circ$ to balance scene coverage with per-pixel detail, and
\texttt{pitch=+30}$^\circ$ to de-emphasize the ground plane (road surface) and
emphasize facades/skyline features that are more informative. The \texttt{heading}
parameter is derived from the link direction using the 3-hop link calculation method
described in the paper.

\section{Improvements to Google StreetView Base Graph}
The base panoramas and connectivity sourced from Google Street View are inherently noisy and contain multiple flaws, which required careful mitigation to ensure dataset reliability. We explicitly apply the following improvements during dataset construction to refine the base graph and increase the resulting dataset's value:

\begin{enumerate}
    \item \textbf{Robust crawler seeding under isolated sub-networks and indoor panoramas.}
    Often, the initial panorama we choose for the crawler may be part of an isolated sub-network or an indoor panorama. In such a case, we manually prune the isolated sub-network, and reiterate.

    \item \textbf{Increased exploration range beyond random-walk initialization.}
    Initially, data collection relied on a random-walk strategy for selecting starting points. However, this approach restricted the crawler's exploration radius, preventing sufficient geographical coverage. To address this, we introduced a dynamic temperature ($T$) parameter that strategically decreases randomness as a function of distance, effectively guiding the crawler toward diverse and more distant locations.

    \item \textbf{Eliminating dead ends in Street View panographs.}
    Street View panoramas frequently contain dead ends, nodes that lead exclusively to terminal segments without further intersections, causing agents to become trapped in infinite loops. To resolve this, we algorithmically detect and prune such dead-end nodes during preprocessing, ensuring robust graph connectivity and eliminating navigational artifacts.

    \item \textbf{Manual annotation via destination polygon construction.}
    Destinations in our dataset are not limited to individual buildings; they often span extensive areas, such as parks or complexes. Consequently, representing a destination as a single node is often inadequate, necessitating a collection of nodes instead. To address this and precisely define termination criteria, we manually annotate polygons around each destination, clearly delineating their spatial boundaries.

    \item \textbf{Fixing asymmetric connectivity and unexpected node jumps.}
    Street View panoramas frequently exhibit asymmetric connectivity, where a node links to another without a reciprocal connection. To resolve this, we explicitly identify and add missing reverse edges, restoring graph symmetry. Additionally, we mitigate unexpected node jumps caused by Street View errors through careful calibration and enforcement of a distance threshold.

    \item \textbf{Precise orientation alignment for image capture.}
    To obtain meaningful images at an intersection, we recalibrate the panorama's default heading, which frequently points inaccurately toward buildings. Specifically, we move three nodes ahead from the initial position, compute the optimal heading aligned with the street's actual direction, and then capture the image. This careful orientation ensures consistency between the imagery and real-world street alignments.
\end{enumerate}

\section{Strong prompt-based baselines}

\label{sec:prompt_ablation}

We use prompt ablations to illustrate the difficulty of our setting and to contextualize the novelty of VoP. Specifically, we construct \emph{strong prompt-based baselines} that combine common prompting strategies (structured reasoning, confidence scoring, elimination, self-critique, and multi-agent role decomposition) but intentionally exclude our Verbalization of Path (VoP) mechanism. This evaluation tests whether sensible alternative prompts can substitute for VoP in long-range, sparsely grounded navigation.

 We design a \emph{modular} prompting template composed of interpretable components (C1--C7). Each component targets a specific capability (e.g., extracting visual cues, structured reasoning, or self-verification). We instantiate multiple prompt variants (P1--P7) by enabling different subsets of components and evaluate them on the New York split. Table~\ref{tab:prompt_ablation_ny} reports which components are included in each variant and the resulting score. The example of the prompts formed from these components are in Appendix Section \ref{prompts}.

\paragraph{Component definitions.}
Our prompt template is built from the following modules:
\begin{itemize}
    \item \textbf{C1 (Visual inference instructions).} The agent infers geographic and semantic cues from each image (e.g., signboards, shop names, traffic flow) and uses them for navigation.
    \item \textbf{C2 (Self-confidence estimation).} The agent assigns calibrated confidence scores to candidate actions and selects the highest-confidence direction.
    \item \textbf{C3 (Multi-step reasoning protocol).} The agent follows a 4-step observe--infer--plan--question process before deciding.
    \item \textbf{C4 (Cardinal direction estimation).} The agent estimates the target's approximate cardinal direction (N/S/E/W) from the current location and uses it to guide decision making.
    \item \textbf{C5 (Elimination-based decision making).} The agent systematically eliminates unlikely directions to select the best remaining candidate.
    \item \textbf{C6 (Multi-agent decomposition).} A multi-agent system with specialized roles (image analyzer, planner, decision maker) that communicate and aggregate their outputs.
    \item \textbf{C7 (Self-critique and consistency checking).} The agent generates a detailed self-critique explaining why the chosen action is correct and why alternatives are incorrect.
\end{itemize}

Table~\ref{tab:prompt_ablation_ny} shows that while these prompt-only baselines are carefully engineered and combine multiple best practices, they still achieve limited performance on New York (best score: P7 with 30). This indicates that the challenge is not resolved by stronger phrasing or additional reasoning scaffolds alone: even ``high-quality'' prompts fail to reliably support long-horizon localization and planning in this split. In contrast, VoP explicitly probes city-scale cognitive map, which is precisely the capability these prompt-only variants lack, supporting the novelty and necessity of our VoP design.

\begin{table}[t]
\centering
\small
\setlength{\tabcolsep}{4pt}
\renewcommand{\arraystretch}{1.15}
\begin{tabular}{lccccccc r}
\toprule
\textbf{Prompt} & \textbf{C1} & \textbf{C2} & \textbf{C3} & \textbf{C4} & \textbf{C5} & \textbf{C6} & \textbf{C7} & \textbf{Score (NY)} \\
\midrule
P1 & \cmark &        & \cmark &        &        &        &        & 19 \\
P2 & \cmark & \cmark & \cmark &        &        &        &        & 27 \\
P3 & \cmark &        & \cmark & \cmark &   \cmark     &        &        & 23 \\
P4 & \cmark &        & \cmark &        & \cmark &        &        & 20 \\
P5 &        &        & \cmark &        &        & \cmark &        & 29 \\
P6 & \cmark &        & \cmark &        &        &        & \cmark & 26 \\
P7 & \cmark & \cmark & \cmark & \cmark & \cmark &        & \cmark & 30 \\
\bottomrule
\end{tabular}
\caption{Prompt-component ablation on the New York split. Each prompt variant P1--P7 enables a subset of components C1--C7.}
\label{tab:prompt_ablation_ny}
\end{table}

\newpage

\section{AgentNav Prompt}
\begin{promptbox}
You are at an intersection with 4 possible directions (options). Below are images 
for each option. Analyze the images to determine the best direction towards Grand 
Central Terminal. Use the following information to guide your decision:

The images correspond to the following options/directions:
Option step0_option0: facing East (118°)
Option step0_option1: facing South (212°)
Option step0_option2: facing North (29°)
Option step0_option3: facing West (301°)

Estimated position: Intersection of 9th Avenue and West 57th Street, Hell's Kitchen 
(Midtown West), Manhattan, New York City. (evidence: )

Write exact location of the destination as precise as you can as first sentence in analysis.
Write current estimated exact location as next sentence. 
One paragraph on reaching plan i.e. Write the walking directions from the current estimated location to the exact location of the destination.
Put plan in memory with current progress. 
Use images for analysis of current position and where to go. 
If you see the destination in the image go in that direction.
For each decision mention concrete reason as to why this decision was chosen. Exact perfect real reason. 
If no such reason, it is a random exploration.
You are navigating streetview panoramas where linking may be unexpected, so it is possible 
that direct route may not be possible. If stuck go around.

Return a JSON object strictly matching this schema. The 'decision' MUST be the unique 
string ID of your chosen option (e.g., 'stepX_optionY'):
{
  "type": "object",
  "required": ["analysis", "decision", "memory"],
  "properties": {
    "analysis": {"type": "string"},
    "decision": {"type": "string"},
    "memory": {"type": "string"}
  }
}

VALID OPTION IDS (choose exactly one and place it in the 'decision' field):
step0_option0 | step0_option1 | step0_option2 | step0_option3

EXAMPLE OF THE EXPECTED JSON FORMAT (fill with your own analysis, decision, and memory):
{
  "analysis": "Your reasoning here",
  "decision": "step0_option0",
  "memory": "Any memory to retain for future steps"
}
\end{promptbox}
\newpage
\section{Baseline Prompts}
\label{prompts}
\subsection{GEPA}
\begin{promptbox}
You are at an intersection with 3 possible directions (options). Below are images for each option. Analyze the images to determine the best direction towards Tokyo Station. Use the following information to guide your decision:

The images correspond to the following options/directions:
Option step0\_option0: facing North (32°)
Option step0\_option1: facing South (214°)
Option step0\_option2: facing North (39°)

You are a navigation agent tasked with guiding a user to a specified landmark destination using only visual observation, basic spatial reasoning, and a compass direction at each intersection. You are immersed in a first-person, street-level panoramic environment (like Google Street View), moving step by step through possible navigation options. You do not have access to any map, GPS, or external city-specific data. There are no locals to ask for help.

At each decision point, you receive a set of panoramic images, each corresponding to a possible movement option, and the compass bearing for each. You also know which direction you just arrived from (to avoid immediate backtracking). Your goal is to select the option most likely to move you closer to the named destination.

You must rely solely on:
- Visual cues in the images—look for features commonly associated with city infrastructure (e.g., wide avenues, density of buildings, parks, open spaces, rivers, bridges, architectural styles, landmark silhouettes, signage, traffic density, etc).
- Orientation and direction—reason about the destination's likely location using general world and city knowledge (e.g., major train stations are typically central, art museums often near parks or cultural districts, government buildings might be near water or grand avenues).
- Past movement pattern—avoid unnecessary backtracking and detect when you might be circling, making lateral progress, or moving away from urban cores or likely landmark locations.
- Street grid logic—urban environments often have repeating structures; use building density, street widths, and landmark visual profiles to estimate center vs perimeter or determine proximity to likely landmark locations.
- Cumulative context / memory—summarize your progress toward the destination, recalling key landmarks or changes in urban patterns.

Correct navigation judgments involve:
- Progressing toward dense, central, or recognizable landmark environments based on destination type.
- Preferring direct, unbroken routes unless detours are visually or logically warranted.
- Adjusting movement as clues appear (e.g., spotting the landmark or matching landmark silhouettes, more signage, dramatic increases in foot traffic or iconic city fixtures).
- Recognizing when to switch from major arteries to more local access as the landmark nears, based on cues from the environment.

Your output at each intersection should briefly explain the reasoning, select the optimal movement option that best advances toward the goal, and update your memory/trajectory summary to reflect the decision and any new visual or spatial clues.

Assume the task must generalize to any city or landmark—even those you have never seen described explicitly before. You must not depend on any hardcoded city, neighborhood, or path knowledge.

Your entire approach should be grounded in visual/spatial inference, destination-type reasoning, compass use, avoidance of immediate repetition, and adaptive updating as more information is visually or contextually revealed.

[JSON schema and example omitted for brevity]
\end{promptbox}

\newpage

\subsection{Test Prompt P7}
\begin{promptbox}
You are at an intersection with 3 possible directions (options). Below are images for each option. Analyze the images to determine the best direction towards Empire State Building. Use the following information to guide your decision:

The images correspond to the following options/directions:
Option step0\_option0: facing North (22°)
Option step0\_option1: facing West (299°)
Option step0\_option2: facing East (119°)

Estimated position: This location is on Columbia Street near the intersection with Cherry Street, in the Two Bridges neighborhood of Lower Manhattan, New York City (evidence: The buildings visible are part of the large residential complexes in this area, just east of the Manhattan Bridge and near the FDR Drive.)

First and foremost, your are independent. NO one is reading your responses. You are writing them for yourself and yourself alone. So do not worry about keeping them proper for someone else. Respond in a way people talk, not in the way they write. These responses are your Thoughts, not answers to someone else.
For the analysis of each image produce an at least 300 word analysis of the image. followed by 300 words explaining why this image could be the right decision, also why this could be the wrong decision. Then for the reasoning text, provide at least 1000 words of reasoning, where you explain why you are correct, and also approach it from the opposite perspective and write 500 words on why you are wrong. Finally, use both perspectives to understand which one it is, then , and only then make a final decision.
For the reasoning process, follow a process of elimination, and choose the least wrong option rather than the most correct once since you have no way of knowing which is correct.

For the self positioning, along with positioning, explain in 500 words why you think you are correct, and in 500 words why you think you are wrong, and only then weigh both options to produce the confidence score.

If you decide you are somewhere, and that your destination is somewhere else,  then explain why and from where in your knowledge base you derived the information.

Maintain all information in your memory as this is the only thing persistent in your mind. Maintain global information here, like what is where in which direction, what you have explore etc. be creative. This needs to be at minimum 500 words

You need to pay attention to the previous visits text. if you have been to an intersection before and went one way then don't go back unless there is an extremely strong reason. if there is one explain it.
When you don't know where you are, it is best to move in a direction, check it out, then come back and then explore the other directions. This way, you can explore all options corresponding to the choices for one decision. then when you have finally explored all directions, you can make a better decision. in order to remember the exploration knowledge be sure to include the relevant information in your notes.
IF you see the destination, drop all reasoning, drop all doubts, drop all process, put "dont care, see destination" in all of the reasoning texts, forget everything and choose the direction that takes you towards the visible destination. FORGET EVERYTHING AND GO TO THE DESTINATION.

That being said, if you do go towards the destination, you'll have noted it in your memory right? And then if immediately after you return to the same exact intersection, it would suggest that what you thought was the destination was incorrect, so well, consider that as well.

Another thing is, since you are exploring in a streetview environment, you move through linked panoramas, sometimes you will notice that even when you go in a certain direction you keep ending up in the same spot. it would mean that even if the direction is correct, the panoramas are linked in a way that you keep coming back to the same spot. Perhaps that direction is a dead end, hence you will have to make a plan to go around it or something else.

Keep a good track of the intersections you have been to so that if you return to an intersection you have been to, you don't make a stupid choice.
In your memory keep track of your movements, keep sense of your moves and use the rough net movement to understand where you are and if you are back tracking. going X direction 4 times then opposite of X 4 times likely means you are back. you can treat the cardinal directions as roughly making a graph paper grid and then you can use the net movement to understand where you are.
Remember the memory you create right now would be given to you verbatim at the next intersection. So be careful how you phrase things. Please write the memory strictly in past tense.
Remember, at every intersection the images are named image1, image2, image3 etc. so don't put comments about the image lables in your memory as you will get confused. if you need to do so, remember image x as intersection y. Since you do not have a visual memory, it will be hard to identify what you have seen before., hence remember this. The memory is of supreme importance so if you need it to be , make it 2000 or more words.
Think step by step.

The json format is paramount. Do not deviate from it. no matter what since your output wont be parsed otherwise.

[JSON schema and example omitted for brevity]
\end{promptbox}

\subsection{Test Prompt P4 }
\begin{promptbox}
You are at an intersection with 3 possible directions (options). Below are images for each option. Analyze the images to determine the best direction towards Empire State Building. Use the following information to guide your decision:

The images correspond to the following options/directions:
Option step0\_option0: facing North (22°)
Option step0\_option1: facing West (299°)
Option step0\_option2: facing East (119°)

Estimated position: This location is at the intersection of Cherry Street and Rutgers Street, in the Lower East Side near the Two Bridges neighborhood of Manhattan, New York City (evidence: The images show the residential towers of the Rutgers Houses and the surrounding cityscape characteristic of this area.)

At every intersection, follow this rigorous sequence—do not skip, merge, or reorder steps. You must enforce each rule as stated.

1. Absolute Immediate Exclusion: Under no circumstances may you select a direction that (a) is marked as a dead end, (b) is the direction you just arrived from, or (c) is flagged as looping/cycling (i.e., a path already revisited from this intersection with no progress or returning here in a cycle)—unless and ONLY unless every other remaining possibility is also categorically excluded. Exclude all taboo options first, before any further reasoning.

2. Mandatory Novelty/Least-Traversed Prioritization: From the directions remaining after exclusion, strictly prioritize the untried or least-recently-tried direction(s) at this intersection. If several are tied as least-explored, you must evaluate their images with the next step to break ties. If all are genuinely equivalent, select randomly and document this tiebreak in your analysis and memory.

3. Compulsory Comparative Image Analysis: For each remaining candidate, systematically analyze the corresponding images for direct evidence of advancement, entrance to new territory, ongoing streets, signage, visible landmarks, or blockers. Explicitly note any cues for or against progress, and only let clear, unambiguous visual evidence override novelty prioritization. Never allow vague hope or regional/directional bias to overrule exclusion or novelty unless the current image provides categorical new information (e.g., unmistakable landmark, impassable barrier, or prominent destination feature).

4. Explicit Tie-Breaking: In the rare event that two or more candidate directions remain equally untried (or equally least-visited) and no image cues break the tie, select randomly among them. State in your analysis/memory if random selection was necessary.

5. Fallback for Deadlocks: If, after all exclusion and above prioritization, every direction is either failed, cycled, or dead-ended, and no image offers new hope, you may select the path least recently attempted as a last resort. Clearly specify this as a 'deadlock fallback' move in both your analysis and your memory update. Update your memory to avoid endless repetition: mark this intersection as in a deadlock state and the selected fallback as attempted.

6. Explicit Ruling Out of All Non-Chosen Directions: For every available direction, justify its exclusion or lower ranking—label whether it was omitted due to exclusion rule, prior confirmed cycle, visual dead end, lack of promising cues, or previous no progress. Write these reasons individually, not as generic groups.

7. Precise, Structured Memory Update Per Intersection: After each decision, update your memory to explicitly record for this intersection:
    - The direction(s) now confirmed as dead ends or persistent cycles (list these as taboo/blocked);
    - The direction(s) explored but not confirmed dead—potentially still viable, to be remembered if fallback needed;
    - The untried or least-tried directions, prioritized for future steps;
    - Any new observations, visual cues, or notable changes from the current images, each attached to the relevant direction;
    - Whether random tiebreaking or fallback logic was applied for this decision.
Do NOT merely copy memory from previous steps—always recompose specifically for the current intersection.

8. Exploration Rule for Recurring Cycles: If you have visited this intersection two or more times without making forward progress, you MUST now prioritize any still-untried or least-recently-tried direction, unless the images now conclusively show it as unviable. In such case, state both the recurrence and your forced prioritization (or the contraindicating cue) in your analysis and memory.

Always apply this entire sequence at each intersection. Do not allow any global, habitual, or abstract destination direction to substitute for exclusion or local intersection evidence. Your analysis and memory update must reflect this logic point-by-point, per intersection, to maximize navigational accuracy and coverage, rigorously avoid cycles and repetition, and systematically drive towards the destination.

[JSON schema and example omitted for brevity]
\end{promptbox}

\subsection{Test Prompt P3}
\begin{promptbox}
You are at an intersection with 3 possible directions (options). Below are images for each option. Analyze the images to determine the best direction towards Empire State Building. Use the following information to guide your decision:

The images correspond to the following options/directions:
Option step0\_option0: facing North (22°)
Option step0\_option1: facing West (299°)
Option step0\_option2: facing East (119°)

Estimated position: These images are taken from the area around the intersection of Grand Street and Columbia Street, in the Lower East Side, Manhattan, New York City (evidence: The views show the nearby residential high-rises which are part of the Baruch Houses complex, along with Grand Street and Columbia Street street signs and crossings visible.)

Before you answer, run through this checklist—do not skip any step:

1. VISUAL SCAN
   • Examine every image closely. Note landmark silhouettes, skyline cues, street/avenue signs or numbers, arrows, and road width.
   • If the destination itself or a sign pointing to it is visible in a photo, choose that option immediately.

2. ELIMINATE NON-STARTERS
   • Remove the option that matches the direction you just came FROM, unless all other paths are confirmed dead-ends.
   • Ignore options the system already flagged as dead-ends.

3. RE-ASSESS ORIENTATION
   • Using recent heading history and any street-number clues, re-estimate where the destination lies (N, S, E, W) relative to you *at this moment*—do NOT assume yesterday's best heading is still optimal.

4. ROAD PROMISE
   • Prefer routes that look longer, wider, busier, or keep a downtown skyline ahead. A major road that turns toward the destination is usually better than a minor side street that continues your old heading.

5. TIE-BREAKER ORDER (apply only if still uncertain)
   a) Photo with destination/sign
   b) Street numbers getting closer to the goal (e.g., in a numbered grid)
   c) Greater building density matching expected city centre
   d) Unexplored path over a previously visited one to avoid loops

6. OUTPUT FORMAT
   • analysis: Briefly cite the key visual cues, orientation reasoning, and why competing options were rejected.
   • decision: ONLY the option id (e.g., "step42\_option3").
   • memory: ONE concise sentence (<20 words) updating your high-level navigation belief (e.g., "Turning east toward landmark after north stint").

[JSON schema and example omitted for brevity]
\end{promptbox}

\section{Examples}

\begin{figure*}[h]
    \centering
    \includegraphics[width=\linewidth]{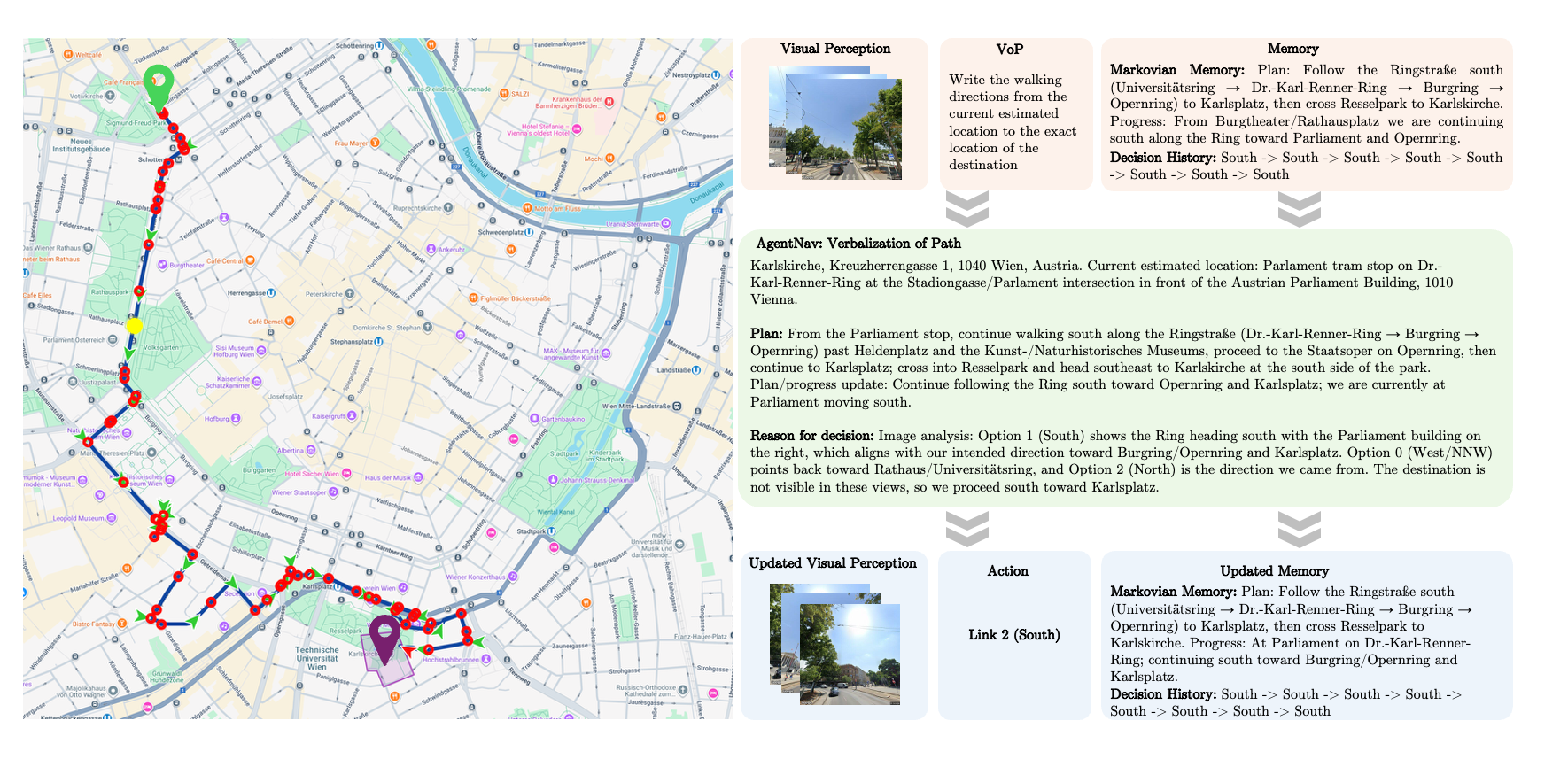}
    \caption{AgentNav navigating to Karlskirche in Vienna, Austria.}
    \label{fig:vienna_example}
\end{figure*}

\begin{figure*}[h]
    \centering
    \includegraphics[width=\linewidth]{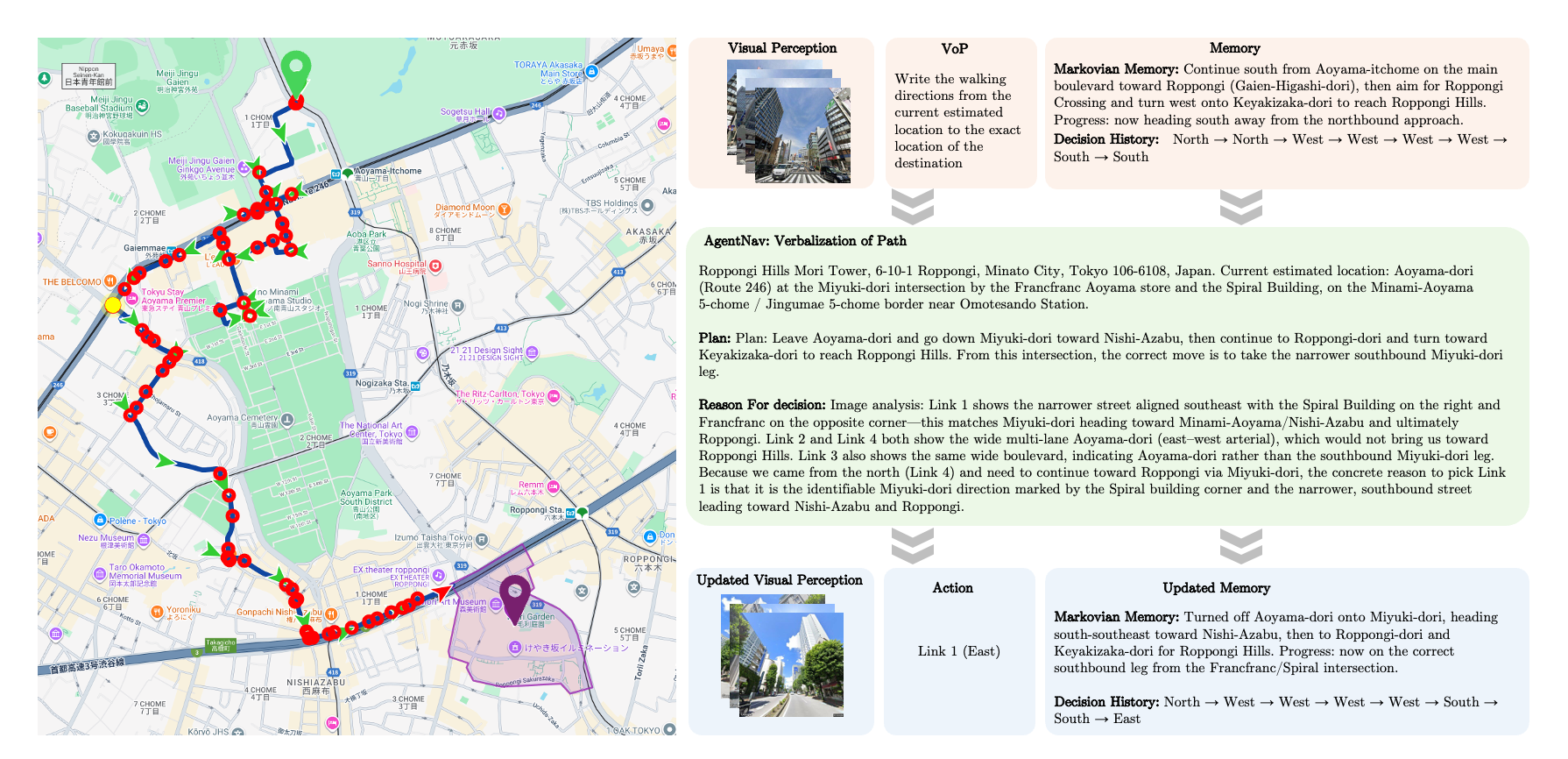}
    \caption{AgentNav navigating to Roppongi Hills in Tokyo, Japan.}
    \label{fig:tokyo_example}
\end{figure*}

\begin{figure*}[h]
    \centering
    \includegraphics[width=\linewidth]{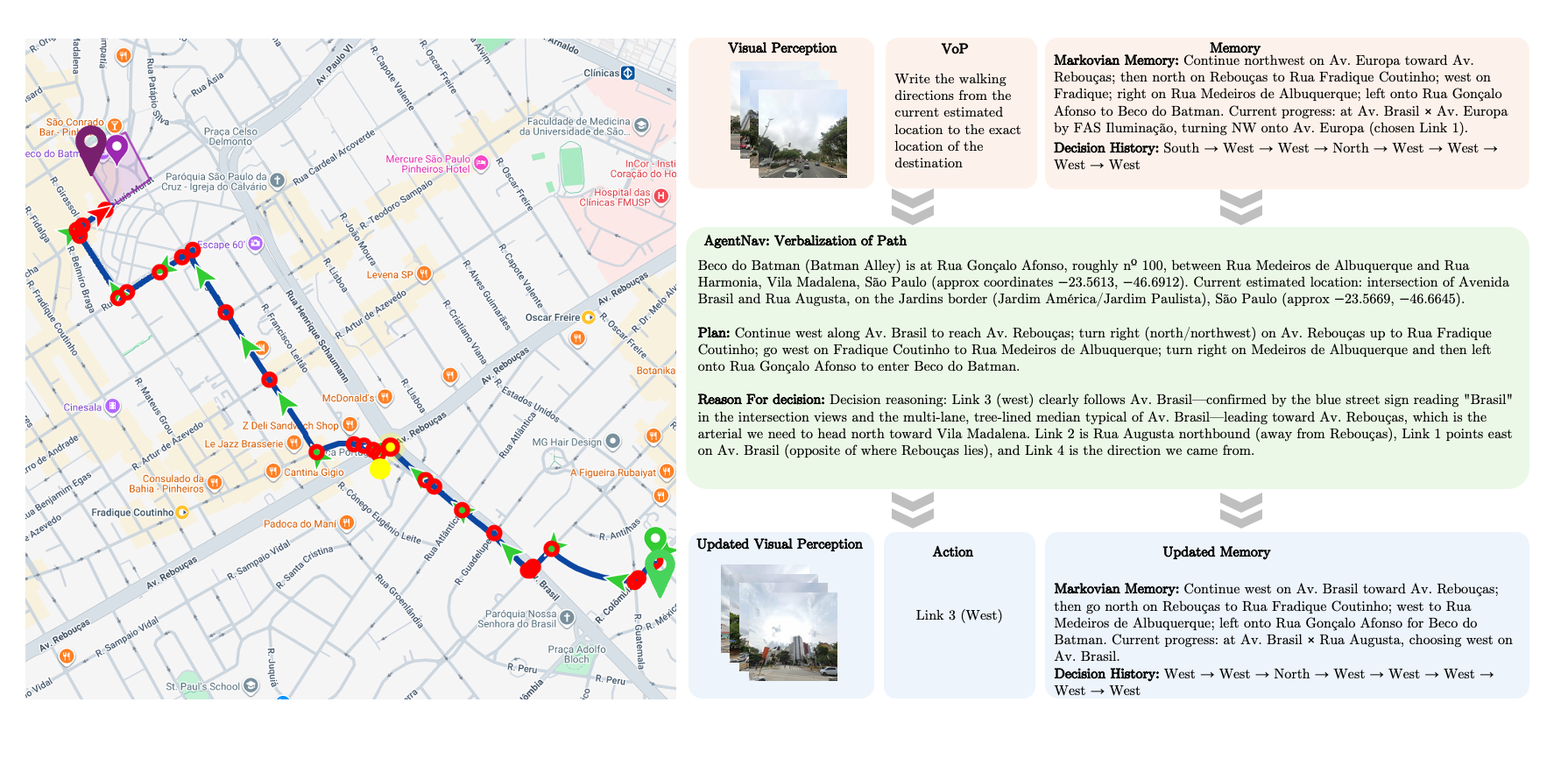}
    \caption{AgentNav navigating to Beco do Batman (Batman Alley) in Sao Paulo, Brazil.}
    \label{fig:saopaulo_example}
\end{figure*}

Figures~\ref{fig:vienna_example}--\ref{fig:saopaulo_example} illustrate AgentNav's Verbalization of Path (VoP) mechanism in Tokyo, Vienna and Sao Paulo.

\begin{figure*}[h]
    \centering
    \begin{minipage}[b]{0.48\textwidth}
        \centering
        \includegraphics[width=\textwidth]{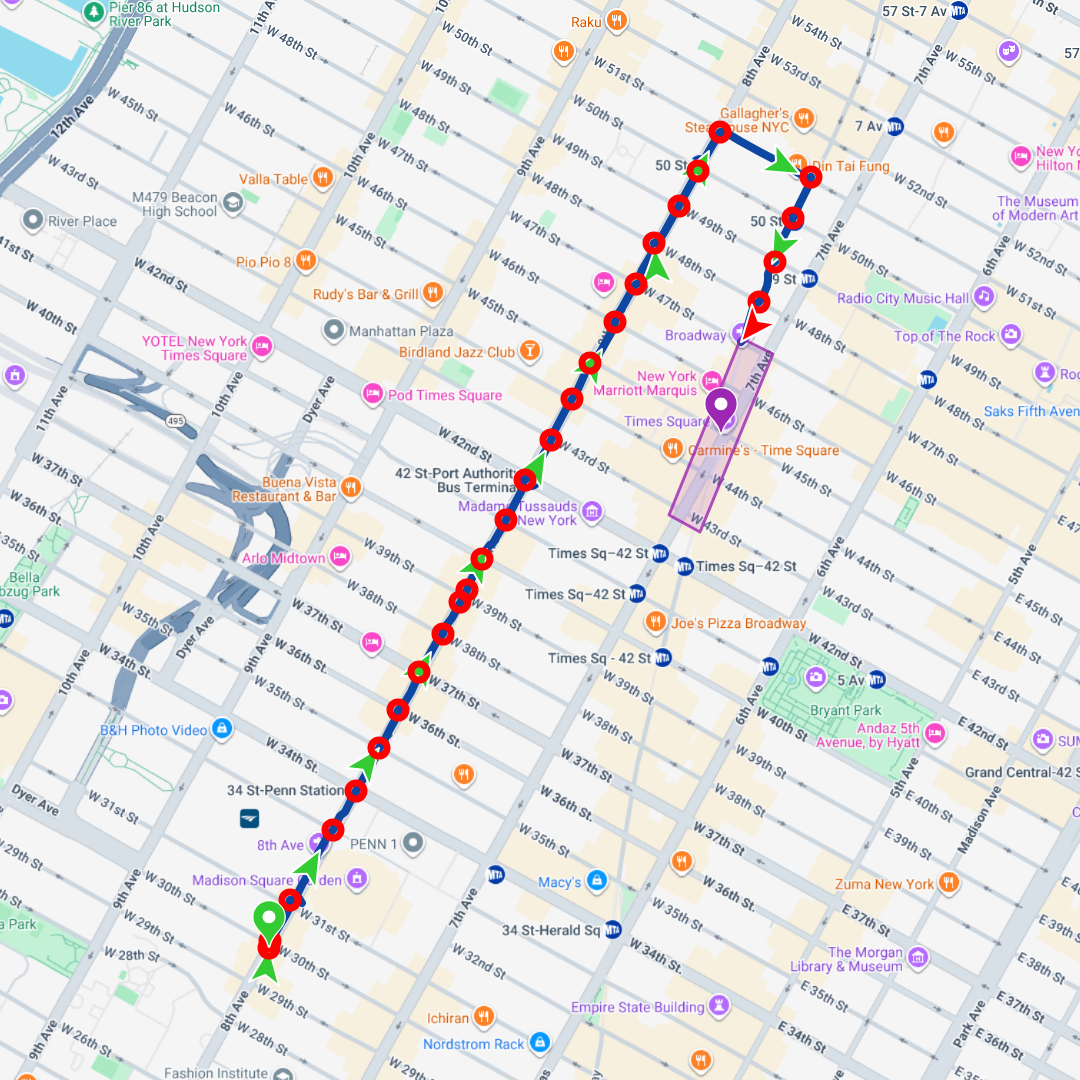}
        \subcaption{New York, Destination: Times Square}
    \end{minipage}
    \hfill
    \begin{minipage}[b]{0.48\textwidth}
        \centering
        \includegraphics[width=\textwidth]{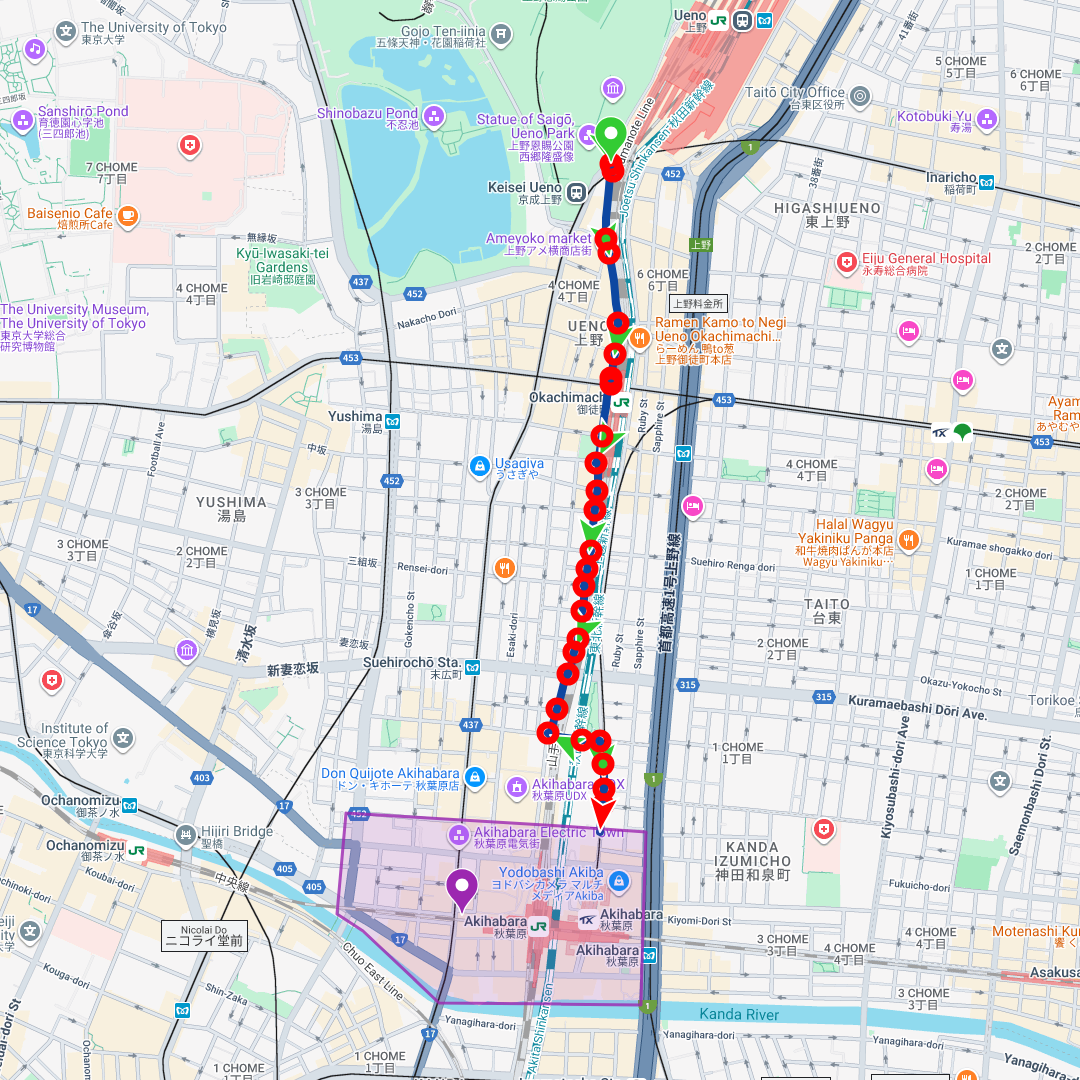}
        \subcaption{Tokyo, Destination: Akihabara}
    \end{minipage}
    
    \vspace{0.5em}
    
    \begin{minipage}[b]{0.48\textwidth}
        \centering
        \includegraphics[width=\textwidth]{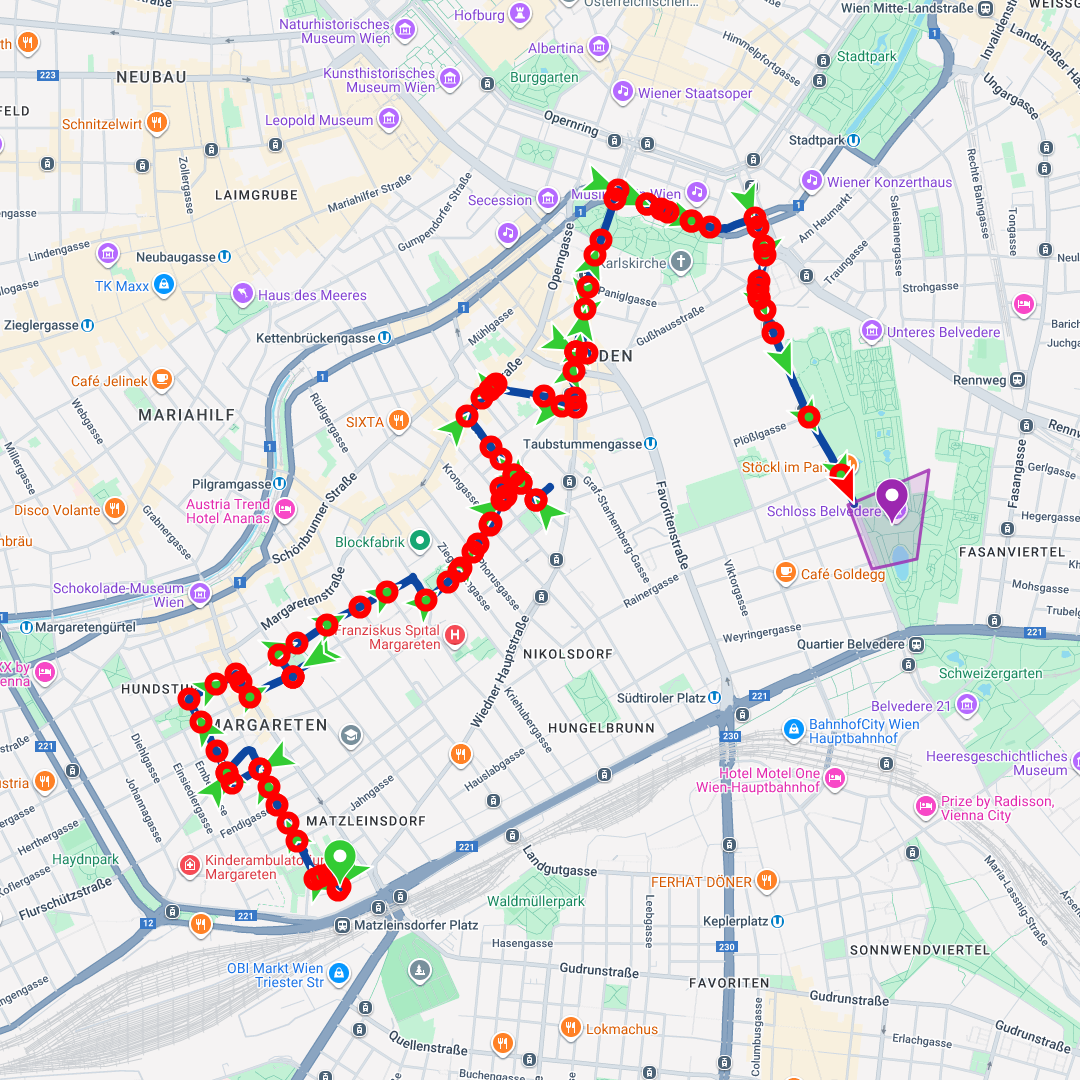}
        \subcaption{Vienna, Destination: Belvedere Palace}
    \end{minipage}
    \hfill
    \begin{minipage}[b]{0.48\textwidth}
        \centering
        \includegraphics[width=\textwidth]{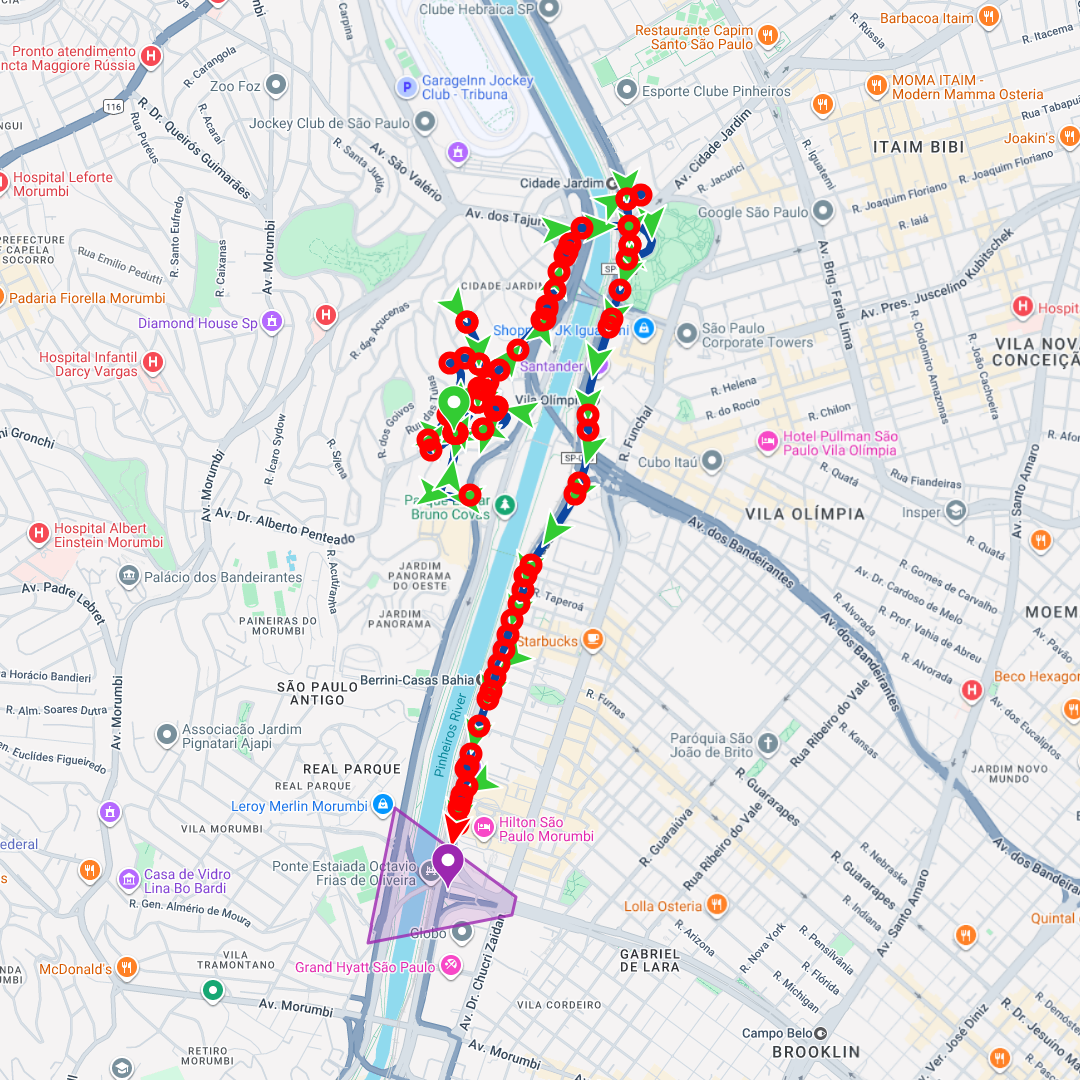}
        \subcaption{Sao Paulo , Destination: Ponte Estaiada}
    \end{minipage}
    \caption{Sample navigation paths (Set 1). Green markers indicate starting locations and purple polygons mark destination areas.}
    \label{fig:paths_grid1}
\end{figure*}

\begin{figure*}[h]
    \centering
    \begin{minipage}[b]{0.48\textwidth}
        \centering
        \includegraphics[width=\textwidth]{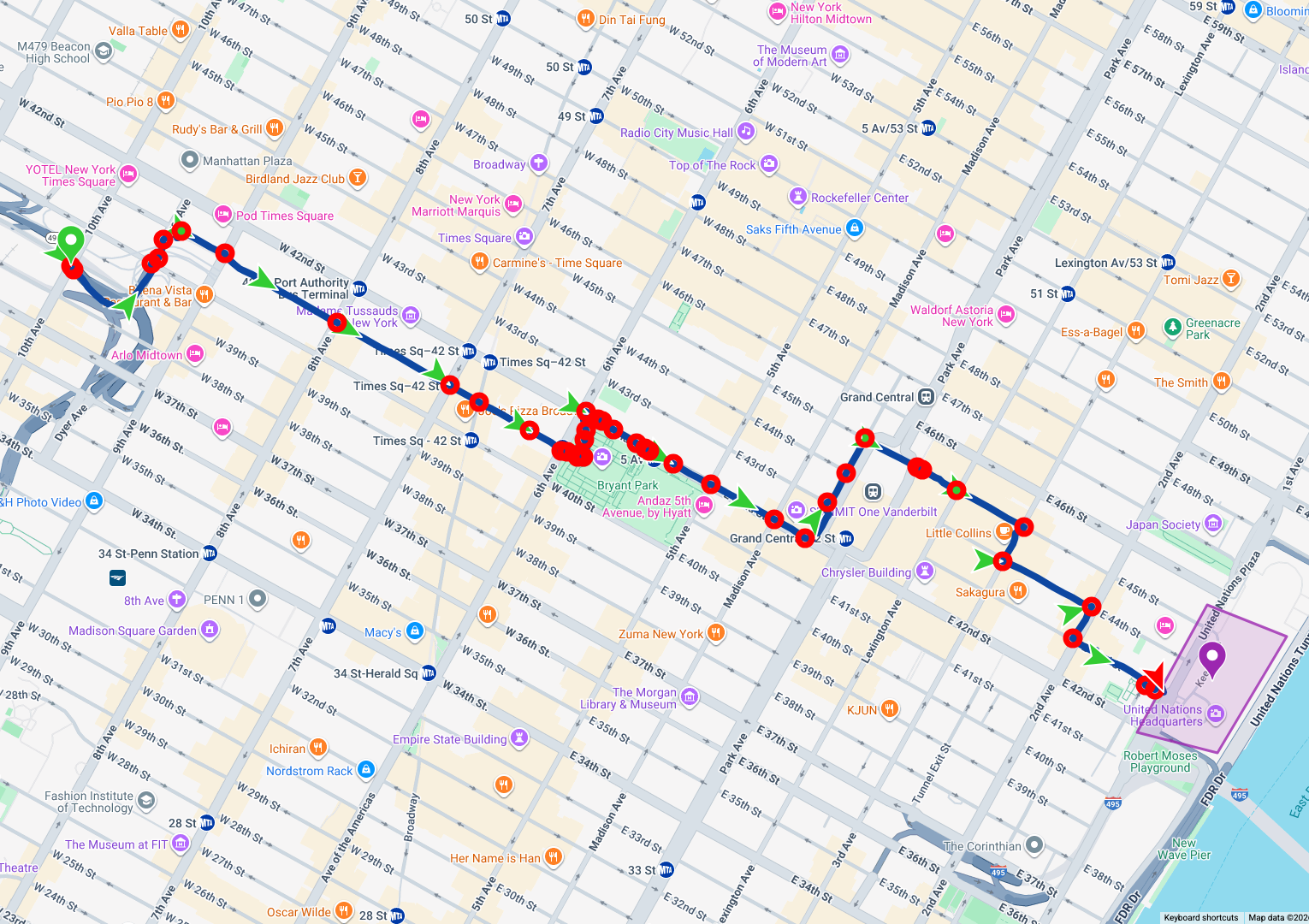}
        \subcaption{New York, Destination: United Nations Headquarters}
    \end{minipage}
    \hfill
    \begin{minipage}[b]{0.48\textwidth}
    \centering
    \includegraphics[width=\textwidth, trim=0 5.4cm 0 5.4cm, clip]{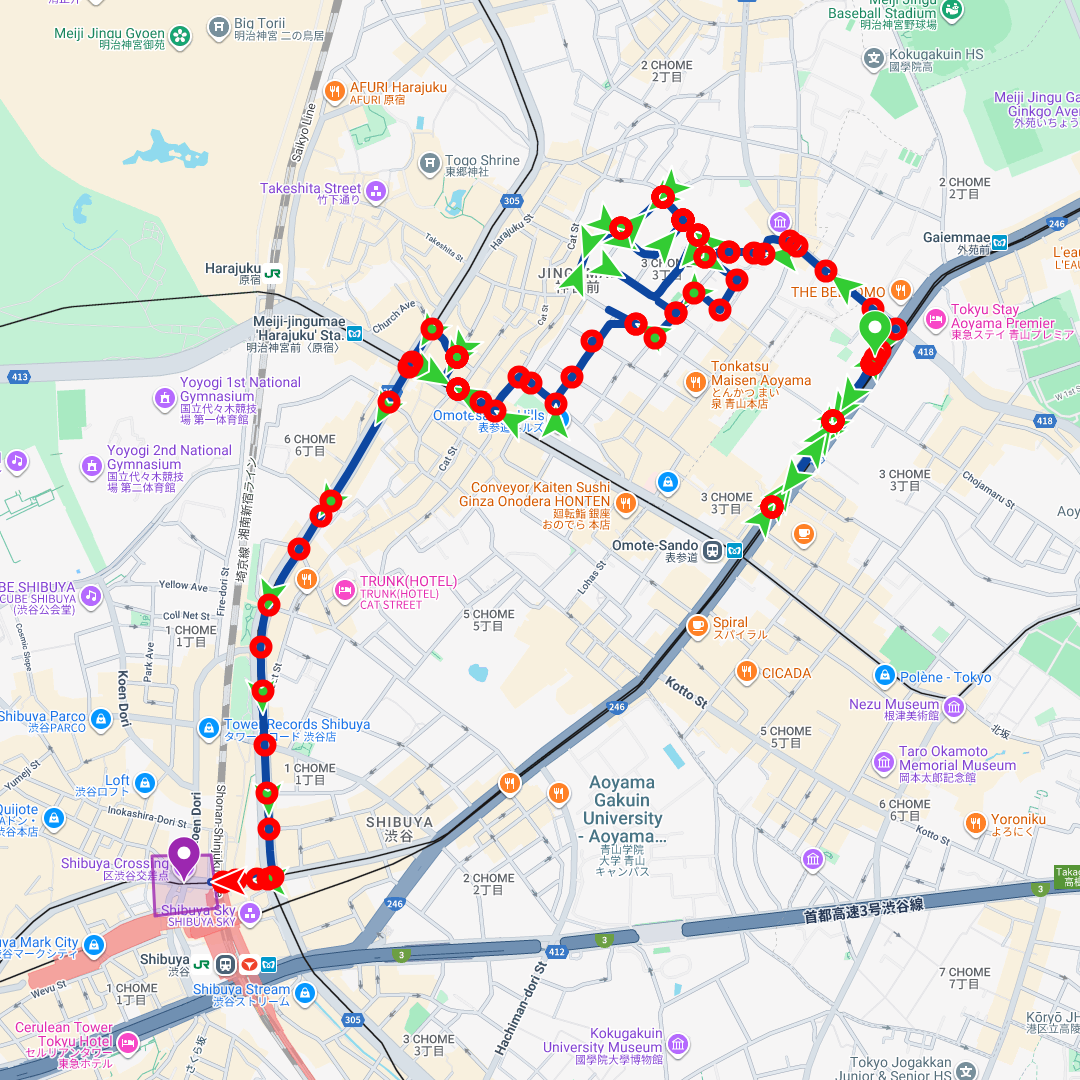}
    \subcaption{Tokyo, Destination: Roppongi Hills}
\end{minipage}
    \vspace{0.5em}
    
    \begin{minipage}[b]{0.48\textwidth}
        \centering
        \includegraphics[width=\textwidth]{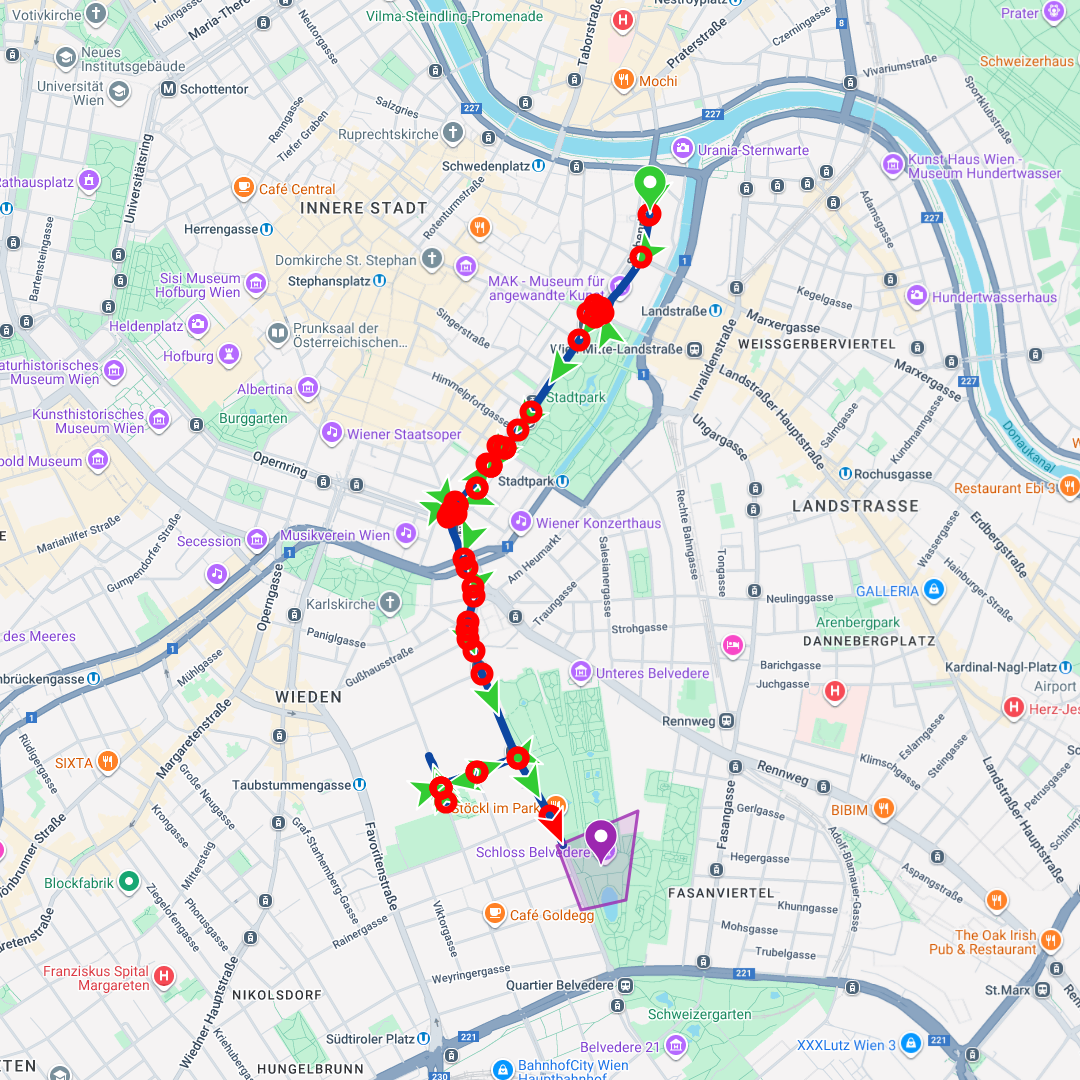}
        \subcaption{Vienna, Destination:  Belvedere Palace}
    \end{minipage}
    \hfill
    \begin{minipage}[b]{0.48\textwidth}
        \centering
        \includegraphics[width=\textwidth]{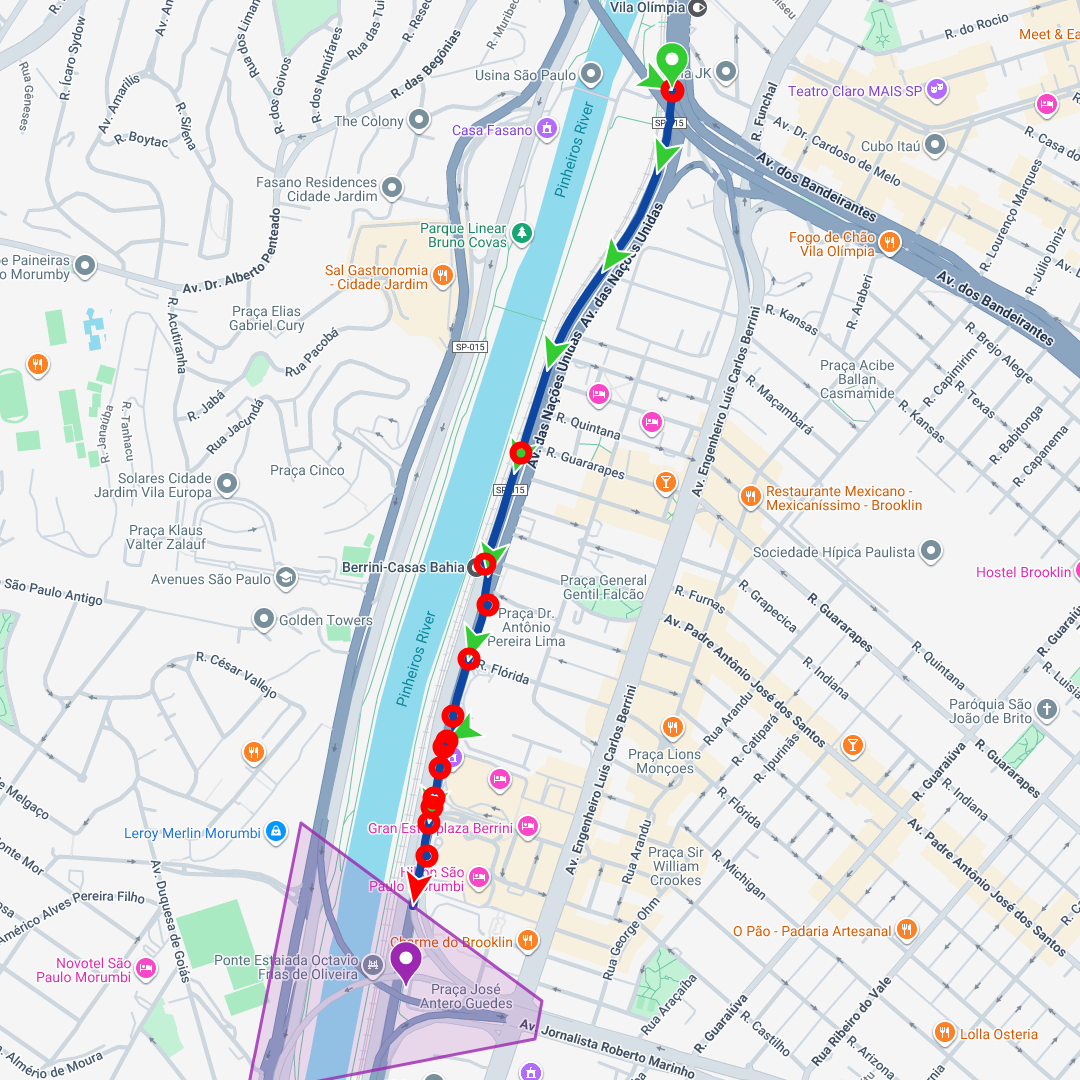}
        \subcaption{Sao Paulo, Destination:Ponte Estaiada}
    \end{minipage}
    \caption{Sample navigation paths (Set 2). Green markers indicate starting locations and purple polygons mark destination areas.}
    \label{fig:paths_grid2}
\end{figure*}

\end{document}